\documentclass[journal]{IEEEtran}

\usepackage{amsmath}					

\usepackage{amssymb}					
\usepackage{amsthm}					
\usepackage{amsfonts}					
\usepackage{mathrsfs}					
\usepackage{mathtools}					
\usepackage{mathdots}					
\usepackage{fancyhdr}					
\usepackage{longtable}					
\usepackage{xcolor}						
\usepackage{multirow, makecell}			
\usepackage{booktabs}					
\usepackage{siunitx}					
\usepackage{nomencl}					
\usepackage[short]{optidef}				
\usepackage[utf8x]{inputenc}				
\usepackage{kantlipsum}					
\usepackage[pdftex]{graphicx} 			
\usepackage{cite}						
\usepackage{pgfplots}					
\usepackage{epstopdf}					
\usepackage{listings}					
\usepackage{enumerate}					
\usepackage{tabularx}					
\usepackage[normalem]{ulem}				
\usepackage{algorithm, algorithmic}		
\usepackage[flushleft]{threeparttable}		
\usepackage{changepage}				

\usepackage{caption}
\usepackage{hyperref}
\usepackage{array}
\usepackage{booktabs} 
\usepackage{enumitem}

\graphicspath{ {/Users/Desktop/} }

\begin{document}
\captionsetup[figure]{name={Fig.},labelsep=period}

\title{Streamlined Federated Unlearning: Unite as One to Be Highly Efficient}

\author{Lei~Zhou, ~Youwen~Zhu, ~Qiao~Xue, ~Ji~Zhang, and ~Pengfei~Zhang

\thanks{-}
}

\markboth{}{}

\maketitle

\begin{abstract}
    Recently, the enactment of ``right to be forgotten" laws and regulations has imposed new privacy requirements on federated learning (FL). Researchers aim to remove the influence of certain data from the trained model without training from scratch through federated unlearning (FU).
    While current FU research has shown progress in enhancing unlearning efficiency, it often results in degraded model performance upon achieving the goal of data unlearning, necessitating additional steps to recover the performance of the unlearned model. Moreover, these approaches also suffer from many shortcomings such as high consumption of computational and storage resources.
    To this end, we propose a streamlined federated unlearning approach (SFU) aimed at effectively removing the influence of the target data while preserving the model performance on the retained data without degradation. We design a practical multi-teacher system that achieves both target data influence removal and model performance preservation by guiding the unlearned model through several distinct teacher models. SFU is both computationally and storage-efficient, highly flexible, and generalizable.
    We conduct extensive experiments on both image and text benchmark datasets. The results demonstrate that SFU significantly improves time and communication efficiency compared to the benchmark retraining method and significantly outperforms existing SOTA methods. Additionally, we verify the effectiveness of SFU using the backdoor attack.
\end{abstract}


\IEEEpeerreviewmaketitle

\section{Introduction}
\IEEEPARstart{T}{he} open sharing of data is of significant importance in driving advancements across various fields, particularly in the rapidly evolving domain of artificial intelligence. Building an effective machine learning model inevitably requires vast amounts of user data for training. However, during data circulation and usage, there exists a substantial risk of privacy leakage \cite{wang2025scu,chundawat2023zero}. Federated learning (FL), as a distributed machine learning approach, was fundamentally designed to enable collaborative training while ensuring privacy protection \cite{mcmahan2017communication}. Nonetheless, the recent enactment of regulations related to the ``right to be forgotten'' has introduced new privacy protection requirements for FL \cite{gdpr2016,klein2020}. When users request the model to ``forget'' a portion of data, simply deleting the targeted data from the dataset is insufficient, as the knowledge derived from this data is already embedded in the local models on each client through prior training.

\begin{figure}[t]
    \centering
    \includegraphics[width=0.4\textwidth]{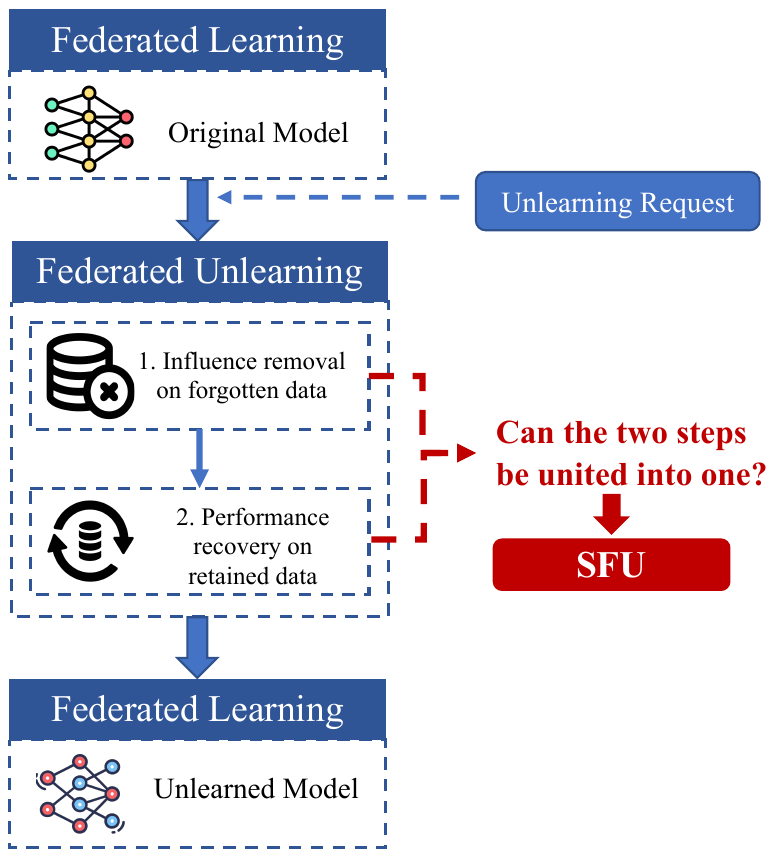} 
    \caption{The general process of federated unlearning.}
    \label{fig1}
\end{figure}

Evidently, the most straightforward approach to addressing this issue is to delete the target data from the dataset and retrain the model. However, this method incurs prohibitive computational costs. Moreover, in the context of FL, retraining may not even be feasible due to the privacy protection requirements of each participating entity. As a result, researchers aim to achieve data unlearning without retraining by utilizing a method known as federated unlearning (FU) \cite{liu2024survey}, which aims to accomplish the objective of forgetting targeted data and producing outcomes similar to those obtained through retraining. This approach seeks to reduce the cost of unlearning substantially while upholding data privacy and security.

FU is designed to efficiently remove specific data and its influence from a trained model without retraining the entire model. 
In contrast to machine unlearning (MU) in centralized environments \cite{bourtoule2021machine,chundawat2023can,wang2023machine, zhang2024forgetting}, FU faces unique challenges in FL environments. This makes many MU methods not directly translatable to the ideas of FU methods, thereby exacerbating the design complexity.
Firstly, the presence of non-independently identically distributed (non-iid) data creates correlations and dependencies among the data, complicating the influence removal of specific data from the global model. Additionally, due to the principle of data privacy protection in FL, data cannot leave the local client, limiting the direct manipulation of raw data during FU and increasing the difficulty of designing effective unlearning mechanisms. Furthermore, compared to MU, FU must complete the unlearning task within a limited number of communication rounds, restricting the execution of complex operations.

In recent years, FU research has made significant progress. As shown in Figure 1, the general workflow of current FU approaches typically involves two main steps: influence removal and performance recovery \cite{jeong2024sok}. Influence removal entails applying unlearning techniques to eliminate the impact of specific target data from the model, aiming to make the behavior of the unlearned model as though it had never encountered the target data. This step has seen a variety of approaches. For instance, gradient manipulation techniques reduce the influence of the target data by adding noise or adjusting gradient directions[6, 34, 70, 80]. However, these approaches require substantial computational resources to determine appropriate noise levels and gradient directions, and the added noise can degrade the model performance on retained data.
Other influence removal approaches include leveraging historical information \cite{cao2023fedrecover,liu2021revfrf,wang2023mitigating,wu2023unlearning,zhang2023fedrecovery}, loss function approximations \cite{jin2023forgettable,liu2022right}, knowledge distillation \cite{li2023federated,wu2023unlearning,dhasade2023quickdrop} and so on. However, these techniques often incur additional storage costs, inaccuracies in estimating data influence, and other limitations. Most critically, the influence removal step in the majority of FU approaches tends to degrade the model performance on non-target data. This necessitates an additional performance recovery step, typically carried out through post-training techniques. Post-training \cite{cao2023fedrecover,gong2022forget,gong2022compressed} enables the model to relearn essential patterns after influence removal but adds both computation time and resource costs. Performance recovery can also be achieved through fine-tuning \cite{dinsdale2022fedharmony,liu2022right,wang2023bfu,zhu2023heterogeneous}, regularization \cite{dhasade2023quickdrop,li2023federated,su2023asynchronous}, and similar techniques, which unavoidably increase the computational burden.

To further improve unlearning efficiency, we considered \textit{can we streamline the steps of FU by combining influence removal and performance recovery into a single step?} To this end, we propose a streamlined federated unlearning method called SFU.
Inspired by knowledge distillation, SFU introduces a multi-teacher system, where multiple distinct teacher models guide the unlearning model (student model) to remove the influence of target data while minimizing performance degradation on the retained data. These teacher models require no additional training, thus avoiding extra computational costs. Furthermore, SFU does not rely on storing historical information or require additional storage resources. SFU demonstrates excellent versatility, as it can be applied to both traditional simple machine learning models and complex deep learning models.
Given the constraints of FL environments, SFU operates without the need for global data access. Instead, each client performs unlearning on its own data, with no restrictions on data distribution. More importantly, SFU offers high flexibility as it does not interrupt the existing FL process. It can seamlessly integrate into any FL training phase when unlearning is required, and after unlearning, the original FL training can resume. For categorical data unlearning, we conducted experiments on image and text benchmark datasets. Results show that SFU significantly improves both time and communication efficiency compared to baseline retraining methods, and far surpasses existing state-of-the-art (SOTA) approaches. Additionally, we implant backdoors on the forgotten data and verify the effectiveness of SFU by comparing the success rate of backdoor attacks before and after unlearning.

The main contributions of this study are as follows:
\begin{enumerate}[] 
    \item We propose a Streamlined Federated Unlearning approach called SFU, which combines the two steps of influence removal and performance recovery into one, dramatically improving unlearning efficiency.
    \item SFU overcomes the shortcomings of existing methods, does not consume additional computational and storage resources, and is applicable to complex deep-learning models.
    \item SFU can be well adapted to FL environments and does not require global data access. Instead, individual clients train on their own data without restrictions on data distribution. Additionally, it offers great flexibility, seamlessly transitioning to any training phase in FL.
    \item Extensive experimental results on image and text datasets show that SFU significantly improves time efficiency and communication efficiency compared to benchmark retraining methods and significantly outperforms existing SOTA methods. Furthermore, we verify the effectiveness of SFU by comparing the success rate of backdoor attacks on embedded backdoors in forgotten data before and after unlearning.
\end{enumerate}

The rest of this paper is organized as follows. In Section \ref{sec:re_work}, we review the relevant literature on federated unlearning. In Section \ref{sec:pre}, we introduce the necessary background knowledge for SFU. Section \ref{sec:approach} provides a detailed description of the proposed SFU method and its specific implementation. Section \ref{sec:theory} demonstrates the theoretical analysis of SFU. Section \ref{sec:experiments} presents the experimental results, and Section \ref{sec:conclusion} concludes the paper.

\section{Related Work}
\label{sec:re_work}
\subsection{Federated Unlearning}
Due to the infeasibility of retraining in FL, current research on FU has explored various directions. 
FedRecover \cite{cao2023fedrecover} tracks the updates from each client during every training round and removes the updates of the target clients during unlearning, using only the retained clients' updates to estimate the unlearned model. Similarly, MetaFul \cite{wang2023mitigating} removes the influence of target data by subtracting the historical updates of the target data from the global model. FedEraser \cite{liu2021federaser} builds on this by employing direction calibration, storing only the most recent updates instead of all historical updates. Other similar methods, such as \cite{jiang2024towards,tao2024communication,yuan2023federated}, rely on historical updates to improve unlearning efficiency to a small extent, but these approaches require large storage resources to maintain historical data.
FFMU \cite{che2023fast} improves robustness to unlearning by adding Gaussian noise to smooth the local model gradients and performs unlearning of the global model during aggregation using methods like the Nemytskii operator. URKL \cite{xu2023revocation} trains noise input to perturb the gradient updates of the target model, facilitating unlearning. In FedFilter \cite{wang2023edge}, the server generates random reverse gradients and performs training using stochastic gradient descent (SGD) to remove the influence of target data. Verifi \cite{gao2024verifi} achieves unlearning of target clients by shrinking the gradients of the target client and amplifying those of the remaining clients. FUMD \cite{zhao2023federated} uses a momentum decay strategy to gradually reduce the model ability to discriminate the target data, effectively removing its influence. FUCDP \cite{wang2022federated} uses TF-IDF to calculate the correlation scores between channels and categories and prunes the channels most similar to the target category to complete unlearning. 
These methods can generally be classified as gradient-based unlearning techniques, which tend to be computationally efficient but are challenging to apply without negatively affecting the overall model performance.
Liu et al. \cite{liu2022right} and Jin et al. \cite{jin2023forgettable} adjust model updates by approximating higher-order derivatives of the loss function, which theoretically offers more refined control but faces significant computational complexity challenges and may lead to performance degradation. 
Reverse training methods, which maximize the loss using gradient ascent instead of descent, are intuitive and effective but significantly reduce model performance and thus require additional performance recovery methods to mitigate the negative impact \cite{halimi2022federated,li2023subspace,wu2022federated}.

\subsection{Knowledge Distillation in FU}
While some existing FU studies have employed knowledge distillation to develop unlearning methodologies, these approaches have been found to have inherent limitations.
FUAF \cite{li2023federated} utilizes a teacher model to generate pseudo-labels with original features for influence removal. However, it still requires elastic weight consolidation for performance recovery. Wu et al. \cite{wu2023unlearning} achieve influence removal of backdoor information by using historical client updates, but this leads to performance degradation. To address this, they employ the original model as a teacher model to guide the unlearned model through the performance recovery step. QuickDrop \cite{dhasade2023quickdrop} generates a representative small synthetic dataset through dataset distillation and then uses this distilled dataset to perform stochastic gradient ascent for effective influence removal of target data. However, it still requires the use of regular stochastic gradient descent (SGD) on the distilled data for performance recovery. 
These methods address either the influence removal or performance recovery step through the teacher-student model, but they still face limitations in terms of unlearning efficiency. The necessity of separate steps for influence removal and performance recovery, along with additional computational overhead, results in suboptimal unlearning efficiency.

\subsection{Discussion}
As shown in table \ref{tab:relatedwork}, the vast majority of existing FU methods lead to a decline in model performance when removing the influence of target data, requiring an additional performance recovery step. Although FU methods based on historical information may avoid performance degradation, their efficiency improvements are minimal, and they require substantial storage resources. FUR \cite{zhang2023fedrecovery} introduces noise and utilizes differential privacy to make the unlearned model indistinguishable from a retrained model. While they claim to eliminate the performance recovery step, the performance of the unlearned model still lags behind that of the retrained model, a gap that cannot be ignored. 
FedU \cite{wang2024fedu} leverages influence functions to estimate and remove the influence of deleted samples from model parameters for unlearning. To counteract performance degradation, they retrain the model on retained data and introduce an adaptive optimization method to balance forgetting and performance retention. However, this balance limits efficiency improvements to a certain extent. Moreover, FedU is specifically designed for unlearning a small number of samples and cannot be easily extended to other unlearning objectives, such as class-level unlearning. To better address these issues, we propose a streamlined FU method called SFU. SFU is based on a specific multi-teacher system, which not only achieves influence removal but also preserves model performance simultaneously. This effectively combines the influence removal and performance recovery steps into one, significantly enhancing unlearning efficiency.

\begin{table}[]
\caption{Summary and comparison}
\label{tab:relatedwork}
\setlength{\tabcolsep}{3pt}
\renewcommand{\arraystretch}{1}
\begin{tabular}{c|cccc}
\toprule[1.5pt] 
Approaches                                                                                & \begin{tabular}[c]{@{}c@{}}Influence\\ removal\end{tabular} & \begin{tabular}[c]{@{}c@{}}High\\ efficiency\end{tabular} & \begin{tabular}[c]{@{}c@{}}Performance\\ maintenance\end{tabular} & \begin{tabular}[c]{@{}c@{}}No fine-tuning\\ required\end{tabular} \\ \hline
\begin{tabular}[c]{@{}c@{}}Storing historical\\ information\\ \cite{cao2023fedrecover,wang2023mitigating,liu2021federaser,jiang2024towards,tao2024communication,yuan2023federated}\end{tabular} & $\checkmark$                                                & $\times$                                                  & $\checkmark$                                                      & $\checkmark$                                                      \\
\begin{tabular}[c]{@{}c@{}}Gradient-based\\ \cite{che2023fast,xu2023revocation,wang2023edge,gao2024verifi,zhao2023federated,wang2022federated}\end{tabular}                      & $\checkmark$                                                & $\checkmark$                                              & $\times$($\downarrow$)                                            & $\times$                                                          \\
\begin{tabular}[c]{@{}c@{}}Loss function\\ approximation\\ \cite{liu2022right,jin2023forgettable}\end{tabular}          & $\checkmark$                                                & $\checkmark$                                              & $\times$($\downarrow$)                                            & $\times$                                                          \\
\begin{tabular}[c]{@{}c@{}}Reverse training\\ \cite{halimi2022federated,li2023subspace,wu2022federated}\end{tabular}                    & $\checkmark$                                                & $\times$                                                  & $\times$($\downarrow$)                                            & $\times$                                                          \\
\begin{tabular}[c]{@{}c@{}}Knowledge distillation\\ \cite{li2023federated,wu2023unlearning,dhasade2023quickdrop}\end{tabular}           & $\checkmark$                                                & $\checkmark$                                              & $\times$($\downarrow$)                                            & $\times$                                                          \\
\begin{tabular}[c]{@{}c@{}}Differential privacy\\ \cite{zhang2023fedrecovery}\end{tabular}                   & $\checkmark$                                                & $\checkmark$                                              & $\times$($\downarrow$)                                            & $\times$                                                          \\
SFU                                                                                       & $\checkmark$                                                & $\checkmark$(highest)                                     & $\checkmark$                                                      & $\checkmark$                                                      \\     \bottomrule[1.5pt] 

\end{tabular}
\end{table}

\section{Preliminary}
\label{sec:pre}
In this section, we introduce the background knowledge and fundamental concepts related to federated learning involved in SFU. 
Basic notations and their meanings used in this paper can be found in Table \ref{tab:symbol}.
\begin{table}[]
    \caption{Basic Notations}
    \label{tab:symbol}
    \setlength{\tabcolsep}{2mm}
    \renewcommand{\arraystretch}{1.25}
    \begin{tabular}{c|c}
    \toprule[1.5pt] 
        Notations      & Descriptions                                              \\ \hline
        $KL$          & Kullback-Leibler divergence                               \\
        $M^P$         & The performance preservation teacher model                \\
        $M^F$         & The forgetting teacher model                              \\
        $M_i^L$       & The label-based preservation teacher model                \\
        $M_i^U$       & Unlearned model on client $i$                           \\
        $M^U$         & Global unlearned Model                                    \\
        $x_i$         & Total data on client $i$                                    \\
        $x_i^f$, $x_i^r$       & Data that needs to be forgotten/retained on client $i$      \\
        $\alpha$           & Hyperparameters                                           \\
        $num(x)$      & Number of samples $x$                                       \\
        $one\_hot$     & One-hot encoding                                          \\
        $y(x)$ & Labels of samples $x$                                       \\
    \bottomrule[1.5pt] 
    \end{tabular}
\end{table}

FL is a decentralized machine learning approach where multiple clients collaboratively train a global model without sharing their private data. Federated averaging (FedAvg) as a common aggregation method is used in various scenarios \cite{mcmahan2017communication}. Each client holds its own local dataset \( x_i \), and the goal is to learn a global model parameter \( M^* \) that minimizes the overall loss across all clients.
In each iteration, the server sends the current global model \( M_t \) to all clients. Each client then performs local training on its dataset using SGD for several epochs, producing an updated model \( M_{i,t+1} \). The server aggregates the model updates from all clients using a weighted average:
\vspace{5pt}
\begin{equation}\label{eqn-1}
    M_{t+1} = \sum_{i=1}^{I} w_i M_{i,t+1}, w_i = \frac{num(x_i)}{\sum_{i=1}^{I} num(x_i)},
    \vspace{5pt}
\end{equation}
where \( w_i \) is the weight assigned to each client based on the size of its dataset, with \( num(x_i) \) being the number of samples $x_i$ in the dataset of the client \( i \).
The objective is to minimize the global loss function:
\vspace{5pt}
\begin{equation}\label{eqn-2}
    M^* = \arg \min_M \frac{1}{I} \sum_{i=1}^{I} w_i L_i(M).
    \vspace{5pt}
\end{equation}

In FedAvg, each client updates its local model and sends the updates to the server, which averages the models to create the new global model. This process continues iteratively, improving the model over time while maintaining data privacy, as clients only share model parameters and not their raw data.

\section{Our Approach}
\label{sec:approach}
In this section, we first introduce the rationale for the teacher-student model and then introduce the design philosophy of each teacher model in the multi-teacher system used in SFU. Then, we provide a detailed description of the overall structure of the system, and finally, we present the specific implementation of SFU.

\subsection{Teacher-student Model}
The concept of the teacher-student model originated from knowledge distillation and is widely used in FL \cite{liu2024spfl,jin2022personalized}. The principle of a general teacher-student model is illustrated below:
\subsubsection{Step 1: Knowledge Extraction}
In the initial phase of the training process, the knowledge base is extracted from the teacher model. Consider a scenario where the teacher model has an output $q$, and the student model has an output $p$. These outputs are typically represented as probability distributions. The raw output of each model is converted into probabilities using the softmax function:
\vspace{5pt}
\begin{equation}\label{eqn-3}
    q=softmax(Z_t/T),
\end{equation}
\begin{equation}\label{eqn-4}
    p=softmax(Z_s/T).
    \vspace{5pt}
\end{equation}
The variables $Z_t$ and $Z_s$ represent the raw outputs of the teacher and student models, respectively. $T$ denotes a hyperparameter known as the distillation temperature, which is used to control the extent of smoothing applied to the probability distribution. Since the objective of this study is to achieve unlearning rather than distillation, the distillation temperature $T$ is set to a constant value of 1.
\subsubsection{Step 2: Knowledge Transfer}
The student model then needs to receive knowledge from the teacher model, thereby reducing the difference between them. In knowledge distillation, the difference between the teacher model and the student model is usually described by KL divergence (Kullback-Leibler divergence). $KL(q \vert\vert p)$ represents the loss of information when the true distribution $q$ of the teacher model is approximated by the distribution $p$ of the student model. In the teacher-student model, the goal is to minimize this difference so that the student model increasingly approximates the teacher model. Thus, the training objective is to minimize:
\vspace{5pt}
\begin{equation}\label{eqn-5}
    KL(q \Vert p)= \sum_k q_k log \frac{q_k}{p_k},
\end{equation}
where $k$ represents the class index, and $q_k$ and $p_k$ are the predicted probabilities of the teacher model and the student model for the $k$-th class, respectively.

When multiple teachers are involved, it can be understood that the student model learns different knowledge from multiple different teacher models simultaneously. The training objective in this scenario is to minimize:
\vspace{5pt}
\begin{equation}\label{eqn-6}
    \begin{aligned}
         & KL(q_1 \Vert p) + \dots +KL(q_j \Vert p) \\
        &= \sum_k q_{1,k} log \frac{q_{1,k}}{p_k} + \dots + \sum_k q_{j,k} log \frac{q_{j,k}}{p_k},
    \end{aligned}
    \vspace{5pt}
\end{equation}
where $q_{1,k}, \dots, q_{j,k}$ and $p_k$ are the predicted probabilities of the $j$ teacher models and the student model for the $k$-th class, respectively.
By minimizing the above loss function, the student model can be made to approach the outputs of multiple teacher models, thus integrating the knowledge from multiple teacher models.

\begin{figure*}[t]
    \centering
    \includegraphics[width=1\textwidth]{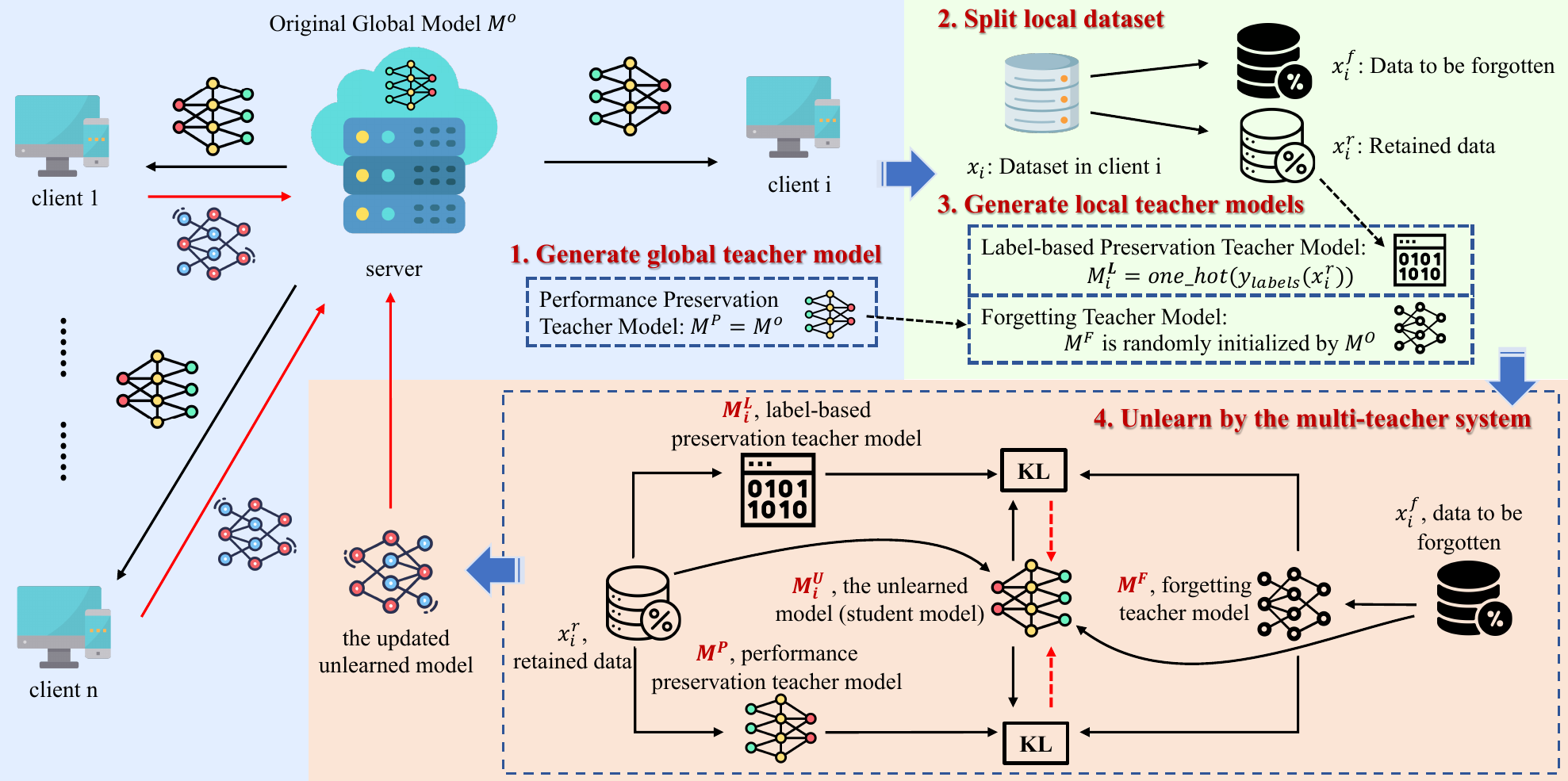} 
    \caption{The framework of SFU. When an unlearning request reaches the server, each client (client $i$, for example) that has the need for the forgotten data carries out the unlearning process in the figure through the multi-teacher system. After completing the unlearning, the corresponding client deletes the forgotten data locally and resumes the normal FL process.}
    \label{fig2}
\end{figure*}

\subsection{Design and Initialization of Teacher Models}
In SFU, three distinct teacher models have been developed to fulfill specific roles and responsibilities, thereby ensuring the efficacy of unlearning and the stability of the model performance. The framework of SFU is shown in Figure \ref{fig2}.
\begin{itemize}
    \item \textbf{Teacher model 1} is designated as the performance preservation teacher model ($M^P$). It is derived from the original model ($M^O$) prior to unlearning. Specifically, the structure and parameters of $M^P$ are identical to those of $M^O$. The objective of $M^P$ is to ensure that the performance of the student model on non-target data remains consistent with that of the original model, thereby preventing any decline in performance. Given that $M^P$ is identical to $M^O$, the original FL process is capable of transmitting both $M^O$ and $M^P$ simultaneously. In other words, the generation of $M^P$ does not require the consumption of additional computational resources.
    \item \textbf{Teacher model 2} is designated as the forgetting teacher model ($M^F$). It is also derived from the original model ($M^O$) prior to unlearning. However, while it shares the same structure as $M^O$, its parameters are randomly initialized to emulate a model that has never encountered the target data. The objective of $M^F$ is to remove the influence of the target data by ensuring that the predictions of the student model on the target data are analogous to those of a random model that has never seen the target data. Since only the structure of $M^O$ is required for generating $M^F$, the initialization of $M^F$ can be conducted by each client when the original FL process transfers $M^O$ to them. Thus, the generation of $M^F$ does not necessitate additional computational resources.
    \item \textbf{Teacher Model 3} is designated as the label-based preservation teacher model ($M_i^L$), which $i$ denotes the $i$-th client. It consists of one-hot encodings of the true labels of the non-target data on each client, making it unique to each client. $M_i^L$ complements $M^P$. Relying solely on $M^P$ is insufficient to eliminate the influence of the target data while preventing a significant decline in model performance. Therefore, we devise $M_i^L$, which aims to align the predictions of the student model on non-target data as closely as possible to the true labels, thereby further preserving the model performance. Since $M_i^L$ is generated based on local data exclusively, no supplementary training is necessary, thereby avoiding additional computational resource consumption.
\end{itemize}

\subsection{Design of the Multi-teacher System}
In accordance with the privacy requirements of FL, the unlearning process in SFU is conducted on each client. We illustrate the unlearning process on a specific client (Client $i$). The student model, or the unlearned model, is initialized to the state of the original model for that client at the time the unlearning request is received. The goal of the unlearned model is to eliminate the influence of the target data while preserving the non-target data by acquiring diverse knowledge from the individual instructor models.

Specifically, assume that the global primitive model is denoted as $M^O$ and that the local primitive model is identical to the global primitive model prior to the commencement of unlearning, which is also $M^O$. The local unlearned model, designated as $M_i^U$, begins the unlearning process based on the local model upon receiving the unlearning command. Consequently, at the inception of unlearning, $M_i^U$ can be defined as:
\begin{equation}\label{eqn-7}
    M_i^U = M^O.
\end{equation}

The local dataset consists of $x_i$, which can be expressed as the sum of $x_i^r$ (the target data to be forgotten) and $x_i^f$ (the non-target data to be retained). According to the previous definition, the performance preservation teacher model is designated as $M^P$, whereas the forgetting teacher model is designated as $M^F$. 
Since the generation of $M^F$ only requires the structure of the global original model $M^O$ and not its parameter distribution, we have:
\vspace{5pt}
\begin{equation}\label{eqn-8}
    M^P = random\_initialize(M^O).
\end{equation}

In light of the aforementioned, it can be posited that the objective of $M_i^U$ is to facilitate the unlearning of the target data $x_i^f$, thereby approximating the predictions of $M^F$. This is to reduce the discrepancy between the performance of $M_i^U$ and $M^F$ on the target data. Additionally, $M_i^U$ strives to maintain the performance on the retained data $x_i^f$ without any deterioration, aligning its predictions closely with the original state, thereby reducing the performance gap between $M_i^U$ and $M^P$ on the non-target data. To quantify the discrepancy between the two models, we have elected to employ the KL-Divergence metric. Consequently, the objective delineated for SFU is to minimize:
\vspace{5pt}
\begin{equation}\label{eqn-9}
    KL(M_i^U(x_i) \parallel (\alpha \cdot M^P(x_i^r ) + M^F(x_i^f))),
    \vspace{5pt}
\end{equation}
where $M_i^u (x_i)$ represents the predictive distribution of the unlearned model on the local client data, $M^F (x_i^f )$ represents the predictive distribution of the forgetting teacher model on the target forgotten data, and $M^P (x_i^r )$ represents the predictive distribution of the performance preservation teacher on the retained data. $\alpha=num(x^r))/(num(x^f)$ is a customized hyperparameter to balance the quantitative gap between the target data and the retained data to safeguard the unlearning performance. The detailed analysis of $\alpha$ can be found in section \ref{sec:experiments}.

Nevertheless, relying on the performance preservation teacher model in isolation is insufficient to achieve the objective of removing the influence of the target data while maintaining the performance of the unlearned model on the retained data. Therefore, to further enhance the ability of SFU performance preservation so as to omit the step of performance recovery, we further design the label-based preservation teacher model ($M_i^L$) on the local client based on the data characteristics of each client. $M_i^L$ is defined as:
\vspace{5pt}
\begin{equation}\label{eqn-10}
    M_i^L= one\_hot(y(x_i^r)),
    \vspace{5pt}
\end{equation}
which represents the $one\_hot$ encoding of the true labels of the retained non-target data on the local client. The objective is to narrow the performance gap between $M_i^U$ and both $M^F$ on $x_i^f$ and $M_i^L$, thereby ensuring the continued reliability of the unlearned model on the retained non-target data. Accordingly, the second objective of SFU is to minimize:
\vspace{5pt}
\begin{equation}\label{eqn-11}
    KL(M_i^U (x_i )) \Vert (M_i^L (x_i^r)+\alpha \cdot M^F (x_i^f )),
    \vspace{5pt}
\end{equation}
where $M_i^t(x_i^r)$ represents the true distribution of labels on the retained non-target data. 

Overall, in order to synthesize the roles of these teacher models and achieve the goal of removing the influence of the forgotten data while maintaining the performance of the model, the total unlearning goal of Client $i$ is summarized as below:
\vspace{5pt}
\begin{equation}\label{eqn-12}
    \begin{aligned}
        &KL(M_i^U (x_i )) \Vert (M^P  (x_i^r )+\alpha \cdot M^F (x_i^f ))\\ 
        &+ KL(M_i^U (x_i )) \Vert (M_i^L (x_i^r)+\alpha \cdot M^F (x_i^f )).
    \end{aligned}
\end{equation}

\subsection{Realization of SFU}
\begin{algorithm}[h]
    \caption{SFU via Multi-Teacher System}
    \label{alhorithm sfu}
    \begin{algorithmic}[1]
    \REQUIRE Original global model $M^O$, the target data to be forgotten $x^f$, the retained data $x^r$.
    \ENSURE Unlearned global model $M^U$.\\
    \textbf{Server Executes:}
    \STATE Issue unlearning instructions and send the global model $M^O$ to all clients.\\
    \textbf{Client Executes:}
    \FORALL{clients in parallel}
        \STATE Divide the local dataset $x_i$ into $x_i^f$ (forgotten data) and $x_i^r$ (retain data).
        \STATE Generate three teacher models: ${M^P} \leftarrow {M^O}$, $M^F \leftarrow random ({M^O})$ and $M_i^L \leftarrow one\_hot(y_{labels}(x_i^r))$.
        \STATE Initial the unlearned (student) model ${M_i^U} \leftarrow {M^O}$.
        \STATE Update $M_i^U$ by the multi-teacher system based on Equation \ref{eqn-12}.
    \ENDFOR \\
    \textbf{Server Executes:}
    \STATE Aggregate $M_i^U$ to generate unlearned global model $M^U$.
    \IF {$M^U$ does not satisfy the forgetting criterion}
        \STATE Repeat steps from 6 to 9 lines.
    \ELSE
        \FORALL{clients in parallel}
            \STATE Delete $x_i^f$ from the local dataset $x_i$ and keep $x_i^r$.
        \ENDFOR
    \ENDIF
    \RETURN $M^U$
    \end{algorithmic}
\end{algorithm}

In the FL environment, the server maintains a global model, while each client retains its local data and a local model. This study is based on the assumption that the environment is isomorphic, which implies that all clients possess the same model architecture. Data are not shared between clients, and the server and local models are updated periodically solely through the exchange of model parameters. 

In the event that a user issues an unlearning request during the standard FL training process, SFU performs unlearning through the multi-teacher system, as shown in Algorithm \ref{alhorithm sfu}. 
Firstly, upon receiving the unlearning request, the server sends an unlearning directive to each client and transmits the current global model ($M^O$), generating and sharing the performance preservation teacher model ($M^P$) with all clients. Subsequently, upon receiving the directive, each client categorizes its local data into the target data to be forgotten ($x_i^f$) and the remaining non-target data ($x_i^r$). Based on the received performance preservation teacher model ($M^P$) and the classified non-target data, each client generates a forgetting teacher model ($M^F$) and a label-based preservation teacher model ($M_i^L$). Using a multi-teacher system, clients locally initiate the SFU unlearning operation and upload the updated model parameters to the server. The server aggregates the updates from all clients and checks whether the unlearning process is complete. If not, it instructs the clients to repeat the aforementioned steps until the unlearning is finalized. Once the unlearning is completed, the server notifies each client to update their local datasets by removing the target data to be forgotten ($x_i^f$) while retaining the remaining non-target data ($x_i^r$), ultimately restoring the system to its original federated learning training process.

\section{Theoretical Analysis}
\label{sec:theory}
This section theoretically analyzes how SFU achieves target data influence removal while preserving the performance of the global model on retained data. Besides, we also analyze the compatibility and flexibility of SFU

\subsection{Theoretical Analysis for Unlearning Effectiveness}
The core objective of SFU is to eliminate the influence of target data \(x_i^f\) while maintaining model utility on non-target data \(x_i^r\). To achieve this, the student model \(M_i^U\) is guided by the multi-teacher system through divergence minimization. Specifically, the optimization objective for target data \(x_i^f\) is dominated by the forgetting teacher \(M^F\), whose predictions approximate a uniform distribution due to random initialization.
The divergence term for target data is formulated as:
\vspace{5pt}
\begin{equation}\label{eqn-13}
    \begin{aligned}
        \mathcal{D}_{\text{forget}} = \sum_{x_i^f} M_i^U(x_i^f) \log \frac{M_i^U(x_i^f)}{ M^F(x_i^f)}.
    \end{aligned}
\end{equation}
Minimizing \(\mathcal{D}_{\text{forget}}\) forces the predictions of the student model on \(x_i^f\) to align with \(M^F(x_i^f)\). Since \(M^F\) is untrained on \(x_i^f\), its output distribution \(q_F = M^F(x_i^f)\) is approximately uniform, i.e., \(q_F \propto \frac{1}{K}\) for \(K\) classes. This ensures:
\vspace{5pt}
\begin{equation}\label{eqn-14}
    \begin{aligned}
        \lim_{\mathcal{D}_{\text{forget}} \to 0} M_i^U(x_i^f) \to \text{Uniform}(K),  
    \end{aligned}
\end{equation}
effectively erasing the influence of \(x_i^f\).
To validate this, consider the gradient of \(\mathcal{D}_{\text{forget}}\) with respect to the parameters \(\theta\) of the student model:
\vspace{5pt}
\begin{equation}\label{eqn-15}
    \begin{aligned}
        \nabla_\theta \mathcal{D}_{\text{forget}} = \sum_{x_i^f} \left( \frac{\partial M_i^U(x_i^f)}{\partial \theta} \cdot \left(1 - \frac{M_i^U(x_i^f)}{M^F(x_i^f)}\right) \right).      
    \end{aligned}
\end{equation}
As optimization progresses, the gradient drives \(M_i^U(x_i^f)\) toward \(M^F(x_i^f)\), ensuring convergence to an unlearned state.

\subsection{Theoretical Analysis for Performance Preservation}
In addition to removing the influence of target data \(x_i^f\), SFU ensures that the model performance on non-target data \(x_i^r\) remains intact. This is achieved through the joint guidance of two teacher models: the performance preservation teacher \(M^P\) and the label-based preservation teacher \(M_i^L\).

\subsubsection{Role of the Performance Preservation Teacher (\(M^P\))}
The performance preservation teacher \(M^P\) is a direct copy of the original model \(M^O\). Its purpose is to ensure that the student model \(M_i^U\) retains the knowledge of non-target data \(x_i^r\). The divergence term for this objective is defined as:
\vspace{5pt}
\begin{equation}\label{eqn-16}
    \begin{aligned}
        \mathcal{D}_{\text{preserve}} = \sum_{x_i^r} M_i^U(x_i^r) \log \frac{M_i^U(x_i^r)}{M^P(x_i^r)}.  
    \end{aligned}
\end{equation}
Minimizing \(\mathcal{D}_{\text{preserve}}\) ensures that the predictions of the student model on \(x_i^r\) align with those of the original model, i.e., \(M_i^U(x_i^r) \approx M^O(x_i^r)\).
To analyze this, consider the gradient of \(\mathcal{D}_{\text{preserve}}\) with respect to the student model’s parameters \(\theta\):
\vspace{5pt}
\begin{equation}\label{eqn-17}
    \begin{aligned}
        \nabla_\theta \mathcal{D}_{\text{preserve}} = \sum_{x_i^r} \left( \frac{\partial M_i^U(x_i^r)}{\partial \theta} \cdot \left(1 - \frac{M_i^U(x_i^r)}{M^P(x_i^r)}\right) \right).  
    \end{aligned}
\end{equation}
As optimization progresses, the gradient drives \(M_i^U(x_i^r)\) toward \(M^P(x_i^r)\), ensuring that the student model retains the original model performance on non-target data.

\subsubsection{Role of the Label-based Preservation Teacher (\(M_i^L\))}
The label-based preservation teacher \(M_i^L\) provides direct supervision through the ground-truth labels \(y(x_i^r)\), encoded as one-hot vectors. The divergence term for this objective is:
\vspace{5pt}
\begin{equation}\label{eqn-18}
    \begin{aligned}
        \mathcal{D}_{\text{label}} = \sum_{x_i^r} M_i^U(x_i^r) \log \frac{M_i^U(x_i^r)}{y(x_i^r)}.  
    \end{aligned}
\end{equation}
Minimizing \(\mathcal{D}_{\text{label}}\) ensures that the student model’s predictions on \(x_i^r\) align with the true labels, even after unlearning.
The gradient of \(\mathcal{D}_{\text{label}}\) with respect to \(\theta\) is:
\vspace{5pt}
\begin{equation}\label{eqn-19}
    \begin{aligned}
        \nabla_\theta \mathcal{D}_{\text{label}} = \sum_{x_i^r} \left( \frac{\partial M_i^U(x_i^r)}{\partial \theta} \cdot \left(1 - \frac{M_i^U(x_i^r)}{y(x_i^r)}\right) \right).  
    \end{aligned}
\end{equation}
This gradient drives \(M_i^U(x_i^r)\) toward \(y(x_i^r)\), ensuring accurate predictions on non-target data.

\subsection{Compatibility and Flexibility of SFU}
\subsubsection{Non-Disruptive Workflow Integration}
The unlearning process in SFU can be initiated at any stage of the FL training process, whether in the early, intermediate, or near-completion stages. After unlearning is completed, SFU seamlessly resumes the original FL training process without requiring additional steps or modifications to the server-side logic. This non-disruptive design ensures that the FL workflow remains uninterrupted, offering significant practical advantages.
The global model \(M_g\) at round \(t\) is updated as follows:
\begin{enumerate}[label=\alph*)]
  \item \textbf{Normal FL Training (Before Unlearning):}
      \begin{equation}\label{eqn-20}
        \begin{aligned}
                M_g^{t+1} = \frac{1}{N} \sum_{i=1}^N M_i^{t},  
        \end{aligned}
        \end{equation}
    where \(M_i^{t}\) is the local model of client \(i\) at round \(t\).
  \item \textbf{Unlearning Phase:}\\
    During unlearning, the local model \(M_i^{t}\) is updated using the multi-teacher system for r rounds:
        \begin{equation}\label{eqn-21}
        \begin{aligned}
            M_i^{t+r} = M_i^{t} - \eta \sum_{k=1}^{r} \nabla_\theta \mathcal{D}_{\text{total}}^{(k)}.
        \end{aligned}
        \end{equation}
    where \(\mathcal{D}_{\text{total}}\) is the combined divergence term.
  \item \textbf{Resuming FL Training (After Unlearning):}\\
    Once unlearning is complete, the global model resumes normal FL training:
        \begin{equation}\label{eqn-22}
        \begin{aligned}
            M_g^{t+r+1} = \frac{1}{N} \sum_{i=1}^N M_i^{t+r}.  
        \end{aligned}
        \end{equation}
\end{enumerate}

This formulation demonstrates that SFU integrates unlearning into the FL workflow without requiring additional rounds or modifications to the aggregation logic.

\subsubsection{Privacy Preservation and Robustness}
In SFU, the server only receives model updates from clients, which are highly compressed representations of local data. Since the unlearning process is performed locally, no raw data is transmitted to the server. This design preserves data privacy and complies with privacy-centric principles in FL.
Moreover, SFU is robust to non-iid (independent and identically distributed) data distributions across clients, as each client performs unlearning independently based on its local data partition. Let \(x_i = \{x_i^r, x_i^f\}\) denote the local data partition of client \(i\), where \(x_i^r\) is the retained data and \(x_i^f\) is the target data to be forgotten. The unlearning process for client \(i\) depends only on \(x_i\), making SFU invariant to data distribution skewness across clients.

\subsubsection{Computational and Storage Efficiency}
The unlearning process of SFU does not require storing any intermediate information. The three teacher models (\(M^P\), \(M^F\), and \(M_i^L\)) are generated on-the-fly using lightweight operations: \(M^P\) is a direct copy of the original model \(M^O\), \(M^F\) is initialized randomly using the architecture of \(M^O\) and \(M_i^L\) is derived from local labels via one-hot encoding.
While the introduction of teacher models incurs additional computational overhead during local unlearning, SFU requires only a small number of training rounds to complete the process. Compared to current FU methods, SFU achieves significant efficiency gains.

\section{Experiments and Analysis}
\label{sec:experiments}
In this section, we first introduce the relevant settings for the experiments. Then, we conduct detailed experiments for SFU and SOTA approaches from the perspectives of efficacy, fidelity, and efficiency to evaluate the unlearning effect of SFU on both image and text datasets. Additionally, we conduct ablation experiments for combinations of teacher models. Finally, we also experimentally analyze the effect of hyperparameters and supplement the analysis with the effect of unlearning on multiple classes of data. The experimental results and corresponding analyses are presented in each subsection.
\begin{table*}[]
    \caption{Comparison of the unlearning performance of different unlearning methods. Id of target class represents the label of the target data class that needs to be forgotten on each dataset. Acc on all data represents the prediction accuracy of the model before unlearning. Acc on forgotten/retained data represents the prediction accuracy of the model on target forgotten data and remaining data after unlearning, respectively. BD\_ASR represents the success rate of the model by backdoor attack after the target forgotten data is implanted with the backdoor.}
    \label{tab:total}
    \centering
    \setlength{\tabcolsep}{2mm} 
    \renewcommand{\arraystretch}{1.5}
    \begin{tabular}{ccccclccclccclc}
        \toprule[1.5pt] 
\multicolumn{1}{c|}{\multirow{3}{*}{\begin{tabular}[c]{@{}c@{}}Id of \\ target class\end{tabular}}} & \multicolumn{2}{c|}{Before Unlearning} & \multicolumn{12}{c}{After Unlearning} \\ \cline{2-15} 
\multicolumn{1}{c|}{} & \multirow{2}{*}{\begin{tabular}[c]{@{}c@{}}Acc on \\ all data\end{tabular}} & \multicolumn{1}{c|}{\multirow{2}{*}{BD\_ASR}} & \multicolumn{4}{c|}{Acc on forgotten data} & \multicolumn{4}{c|}{Acc on retained data} & \multicolumn{4}{c}{BD\_ASR on forgotten data} \\ \cline{4-15} 
\multicolumn{1}{c|}{} &  & \multicolumn{1}{c|}{} & Retrain & FUCDP & FUR & \multicolumn{1}{c|}{SFU} & Retrain & FUCDP & FUR & \multicolumn{1}{c|}{SFU} & Retrain & FUCDP & FUR & SFU \\ \hline
\multicolumn{15}{c}{CIFAR-10} \\ \hline
\multicolumn{1}{c|}{0} & 84.13 & \multicolumn{1}{c|}{85.52} & 0.07 & 0.89 & 0.72 & \multicolumn{1}{c|}{0.67} & 83.70 & 83.26 & 71.2 & \multicolumn{1}{c|}{83.28} & 0.26 & 2.83 & 2.02 & 0.91 \\
\multicolumn{1}{c|}{5} & 83.92 & \multicolumn{1}{c|}{86.56} & 0.12 & 0.44 & 0.32 & \multicolumn{1}{c|}{0.11} & 84.25 & 83.04 & 72.4 & \multicolumn{1}{c|}{84.81} & 0.53 & 2.66 & 1.87 & 0.57 \\
\multicolumn{1}{c|}{9} & 84.88 & \multicolumn{1}{c|}{85.48} & 0.09 & 0.89 & 0.54 & \multicolumn{1}{c|}{0.44} & 84.26 & 83.58 & 71.7 & \multicolumn{1}{c|}{84.54} & 0.42 & 3.71 & 1.94 & 0.82 \\ \hline
\multicolumn{15}{c}{CIFAR-100} \\ \hline
\multicolumn{1}{c|}{0} & 64.24 & \multicolumn{1}{c|}{83.15} & 0.01 & 0.08 & 0.16 & \multicolumn{1}{c|}{0.12} & 64.95 & 63.08 & 56.32 & \multicolumn{1}{c|}{63.96} & 0.12 & 1.39 & 1.32 & 0.96 \\
\multicolumn{1}{c|}{55} & 64.87 & \multicolumn{1}{c|}{82.80} & 0.05 & 0.54 & 0.49 & \multicolumn{1}{c|}{0.46} & 64.93 & 63.46 & 55.53 & \multicolumn{1}{c|}{65.53} & 0.47 & 3.87 & 2.33 & 1.03 \\
\multicolumn{1}{c|}{99} & 63.94 & \multicolumn{1}{c|}{86.06} & 0.04 & 0.67 & 0.54 & \multicolumn{1}{c|}{0.47} & 64.46 & 62.59 & 55.84 & \multicolumn{1}{c|}{65.47} & 0.58 & 2.34 & 2.67 & 1.42 \\ \hline
\multicolumn{15}{c}{Dbsepia} \\ \hline
\multicolumn{1}{c|}{0} & 96.20 & \multicolumn{1}{c|}{90.24} & 0.02 & - & \multicolumn{1}{c}{-} & \multicolumn{1}{c|}{0.04} & 96.89 & - & \multicolumn{1}{c}{-} & \multicolumn{1}{c|}{96.88} & 0.89 & - & \multicolumn{1}{c}{-} & 0.69 \\
\multicolumn{1}{c|}{5} & 96.44 & \multicolumn{1}{c|}{92.36} & 0.05 & - & \multicolumn{1}{c}{-} & \multicolumn{1}{c|}{0.08} & 96.23 & - & \multicolumn{1}{c}{-} & \multicolumn{1}{c|}{96.54} & 0.46 & - & \multicolumn{1}{c}{-} & 0.83 \\
\multicolumn{1}{c|}{9} & 96.17 & \multicolumn{1}{c|}{91.56} & 0.01 & - & \multicolumn{1}{c}{-} & \multicolumn{1}{c|}{0.24} & 96.44 & - & \multicolumn{1}{c}{-} & \multicolumn{1}{c|}{96.14} & 0.36 & - & \multicolumn{1}{c}{-} & 0.45 \\ 
    \bottomrule[1.5pt] 
    \end{tabular}
\end{table*}

\begin{table}[]
    \caption{Comparison of the efficiency of different unlearning methods. Communication rounds represent the average number of interactions required for unlearning. Time speed-up represents the speed-up ratio in time compared to each method of retraining.}
    \label{tab:efficiency}
    \centering
    \setlength{\tabcolsep}{1mm}
    \renewcommand{\arraystretch}{1.5}
    \begin{tabular}{c|cccc|ccc}
        \toprule[1.5pt] 
        \multirow{2}{*}{Dataset} & \multicolumn{4}{c|}{Communication Rounds} & \multicolumn{3}{c}{Time Speed-up} \\ \cline{2-8} 
 & Retrain & FUCDP & FUR & SFU & FUCDP & FUR & SFU \\ \hline
CIFAR-10 & 312.60 & 55.40 & 18.20 & 1.40 & 7.61$\times$ & 20.45$\times$ & 152.40$\times$ \\
CIFAR-100 & 369.44 & 93.82 & 32.25 & 2.20 & 6.32$\times$ & 18.62$\times$ & 44.04$\times$ \\
Dbpedia & 151.48 & - & - & 1.40 & - & - & 43.45$\times$ \\ 
        \bottomrule[1.5pt] 
    \end{tabular}
\end{table}

\subsection{Experiments Settings}

\subsubsection{Datasets and Models}
To prove the versatility of SFU, we evaluate our proposed method on both image datasets: CIFAR-10, CIFAR-100 \cite{krizhevsky2009learning} and text dataset: DBpedia \cite{zhang2015character}. The CIFAR-10 dataset consists of 60,000 32x32 color images, organized into 10 classes with 6,000 images per class. The CIFAR-100 dataset is similar to the CIFAR-10 dataset but has 100 classes with 600 images per class. The DBpedia dataset consists of 560,000 training samples and 70,000 testing samples, organized into 14 classes.
And we selected different network models for training on each dataset: ResNet18 for CIFAR-10, ResNet44 for CIFAR-100 \cite{he2016deep}, and a simple LSTM \cite{hochreiter1997long} for Dbpesia.

\subsubsection{Training Settings}
In the FL setup, we set the number of clients N to 20. In the normal FL training process, we randomly select 25$\%$ of the clients to participate in each round. During unlearning, each client is required to participate and resume after finishing unlearning. The training data is partitioned according to the Dirichlet distribution to generate non-iid data for the FL process \cite{yurochkin2019bayesian}. We adopt FedAvg \cite{mcmahan2017communication} as the aggregation algorithm.
All the experiments are performed on NVIDIA 3090 (24 GB) with Intel Xeon processors. The experiments are implemented in PyTorch 1.12.0. The KL temperature is set to 1 for all the experiments.

\subsubsection{Baseline Methods}
In addition to comparing the basic method of retraining, we also compare it with the SOTA methods FUCDP \cite{wang2022federated} and FUR \cite{zhang2023fedrecovery} for unlearning class data in FU. FUCDP achieves unlearning by computing the TF-IDF scores of different category channels and pruning the channels with high scores for the target forgotten category. While pruning enables fast unlearning, it also significantly degrades model performance, necessitating fine-tuning for recovery. Similarly, FUR removes specific gradient residues and adds Gaussian noise to achieve the unlearning of specific category data. However, this method also leads to a decline in model performance. Although researchers claim that this degradation is negligible, in reality, the impact remains substantial. Therefore, we introduce an additional fine-tuning step in the unlearning process of FUR to restore model performance.

\subsection{Results and Analysis}
The evaluation of unlearning methods is typically conducted in three dimensions: unlearning efficacy, unlearning fidelity, and unlearning efficiency. Unlearning efficacy assesses the extent to which the forgotten data has been successfully unlearned, indicating whether the unlearning process has been fully completed. Unlearning fidelity is used to describe the performance of the model on the retained data after the unlearning process. A good FU method must ensure the continued availability of the model even after unlearning has been completed. Unlearning efficiency refers to the time and communication efficiency of the FU method in completing unlearning. Once the first two evaluation perspectives have been completed, this dimension provides the most direct means of measuring an FU method, as it reflects the fundamental objective of FU. 
We conduct extensive experiments, with selected results presented. It is important to note that FUCDP is only applicable to CNN models, and FUR is difficult to extend to text datasets. Therefore, we employ an LSTM model for training on the DBpedia dataset, demonstrating the broader applicability of SFU. 

\begin{figure*}[t]
    \centering
    \begin{tabular}{cc}
        \begin{minipage}[b]{0.02\linewidth}
            \centering
            \raisebox{1.2\height}{\rotatebox{90}{\text{Test accuracy ( $\%$ )}}}
        \end{minipage}
        &
        \begin{minipage}[b]{0.98\linewidth}
            \centering
            \begin{minipage}[b]{0.32\linewidth}
                \includegraphics[width=\linewidth]{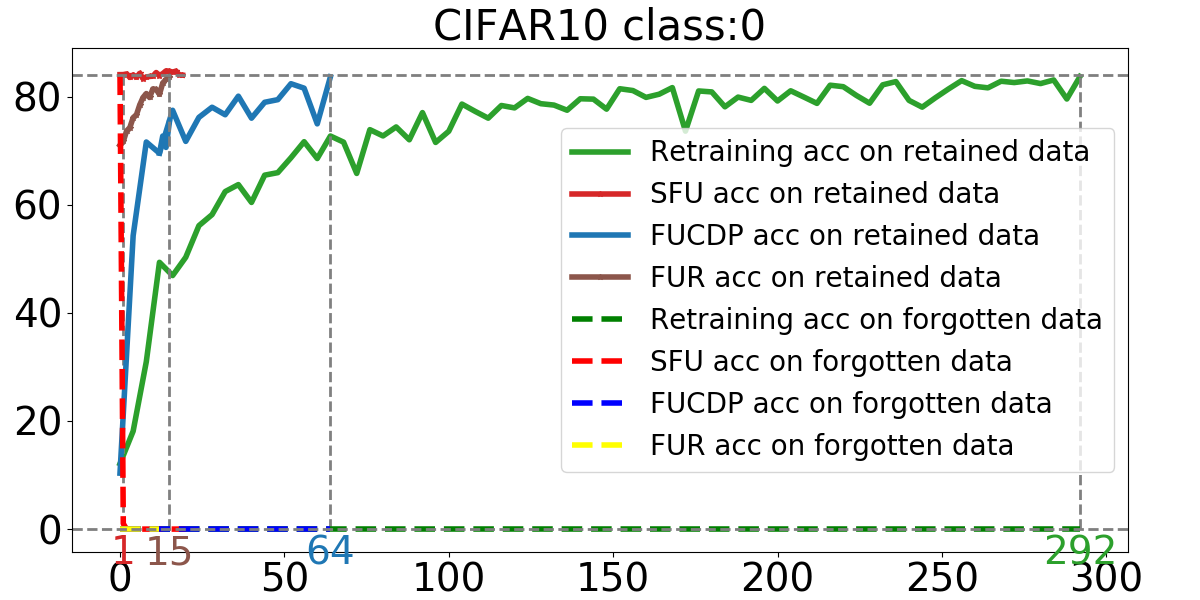}
                \vspace{0.75mm}
                \includegraphics[width=\linewidth]{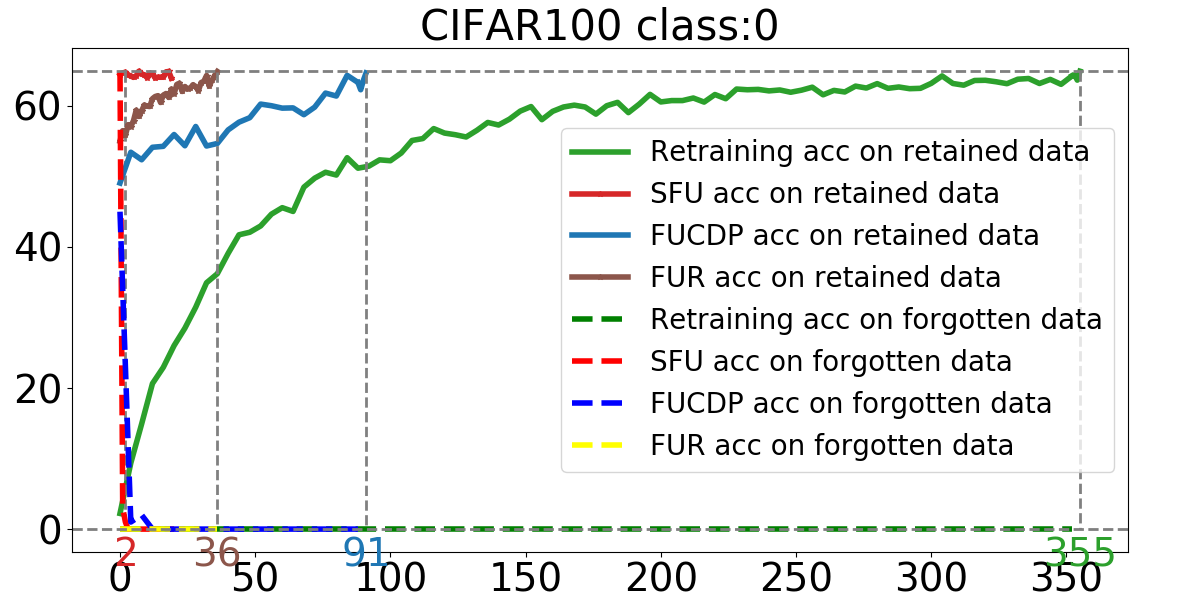}
                \vspace{0.75mm}
                \includegraphics[width=\linewidth]{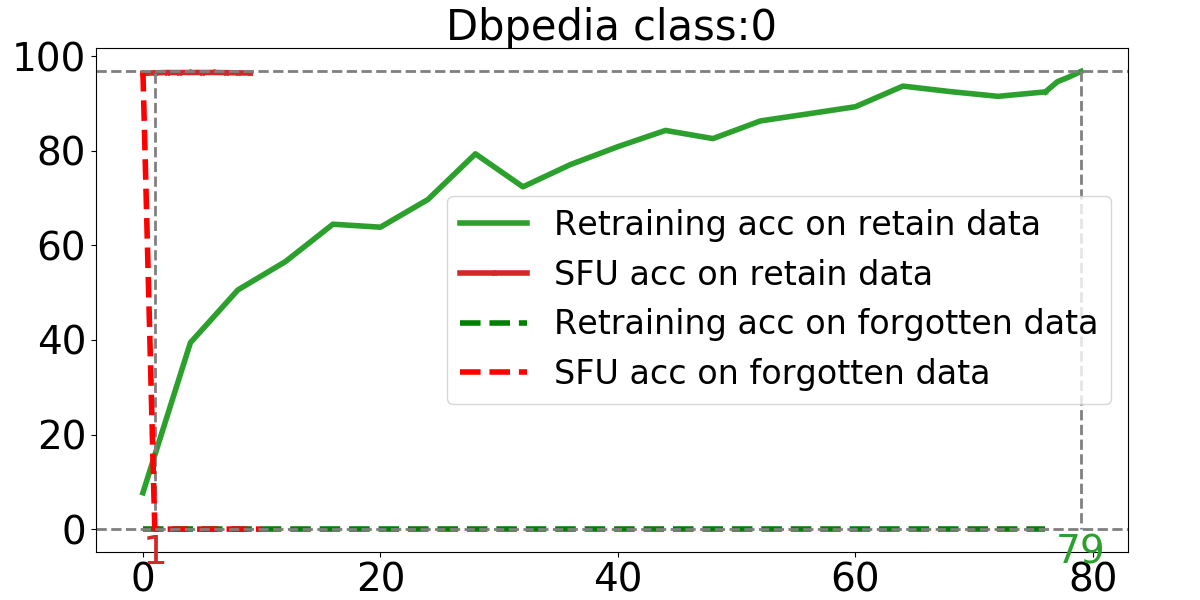}
                \centering Communication rounds
            \end{minipage}
            \begin{minipage}[b]{0.32\linewidth}
                \includegraphics[width=\linewidth]{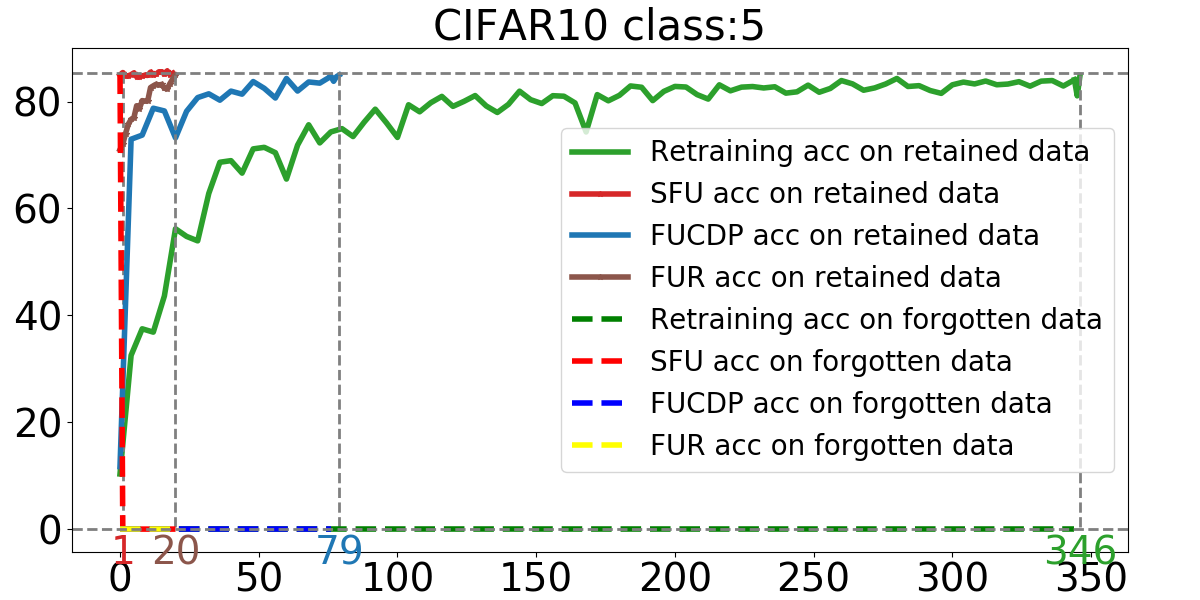}
                \vspace{0.75mm}
                \includegraphics[width=\linewidth]{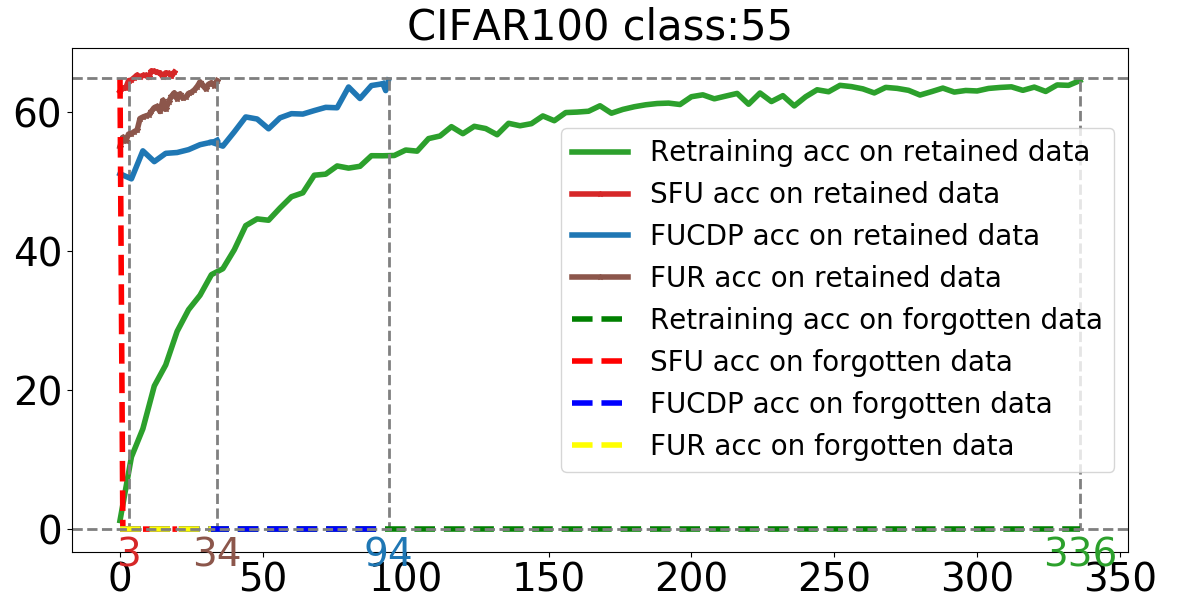}
                \vspace{0.75mm}
                \includegraphics[width=\linewidth]{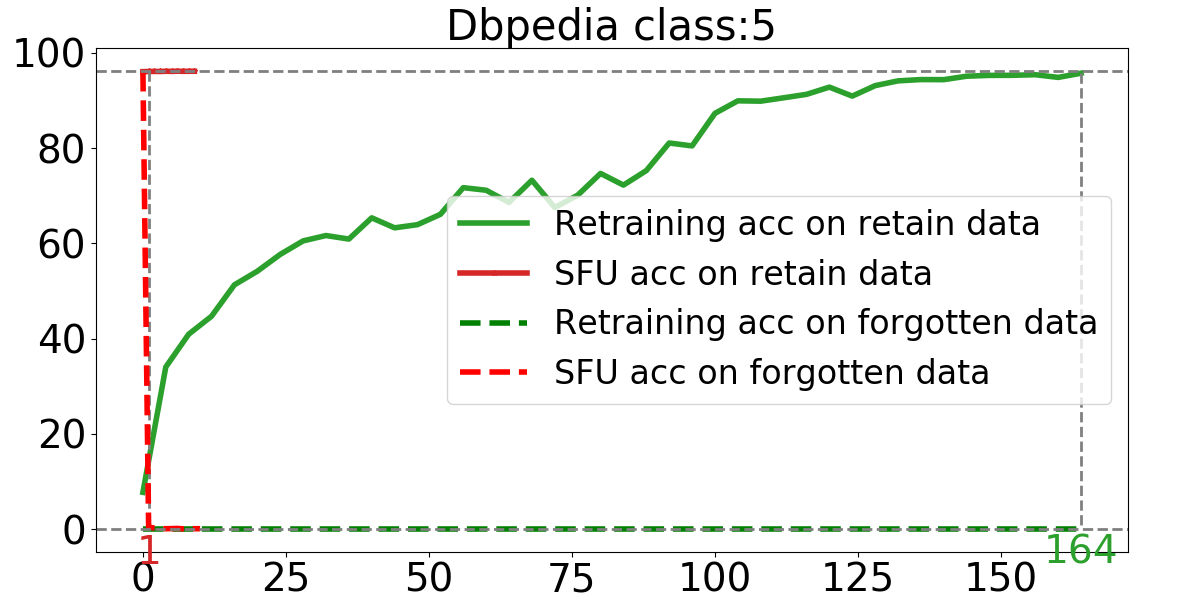}
                \centering Communication rounds
            \end{minipage}
            \begin{minipage}[b]{0.32\linewidth}
                \includegraphics[width=\linewidth]{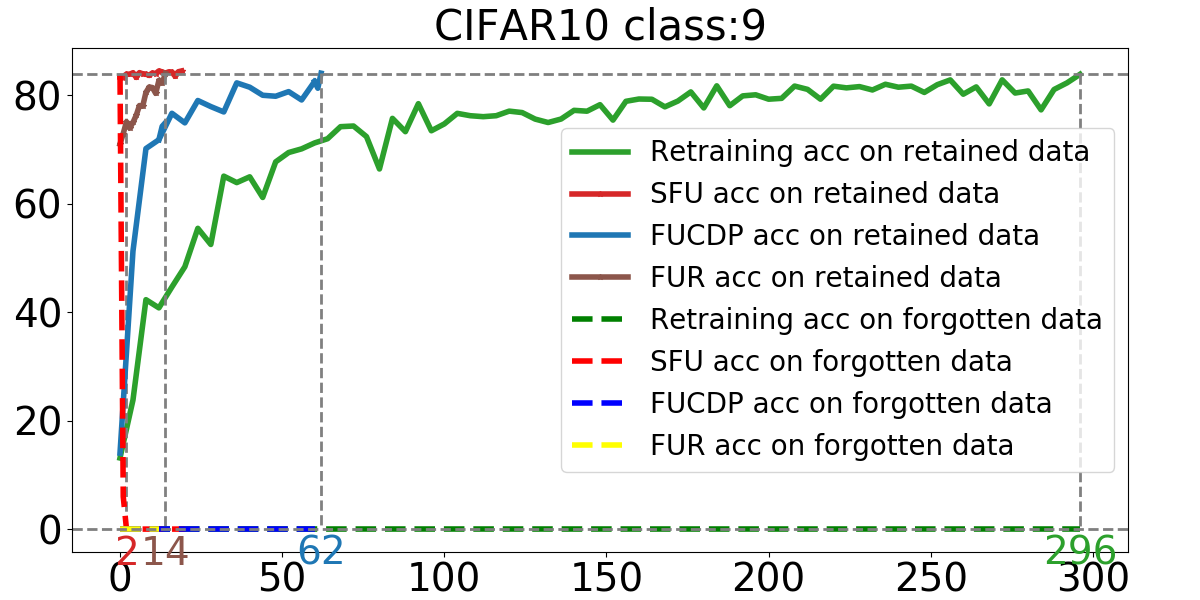}
                \vspace{0.75mm}
                \includegraphics[width=\linewidth]{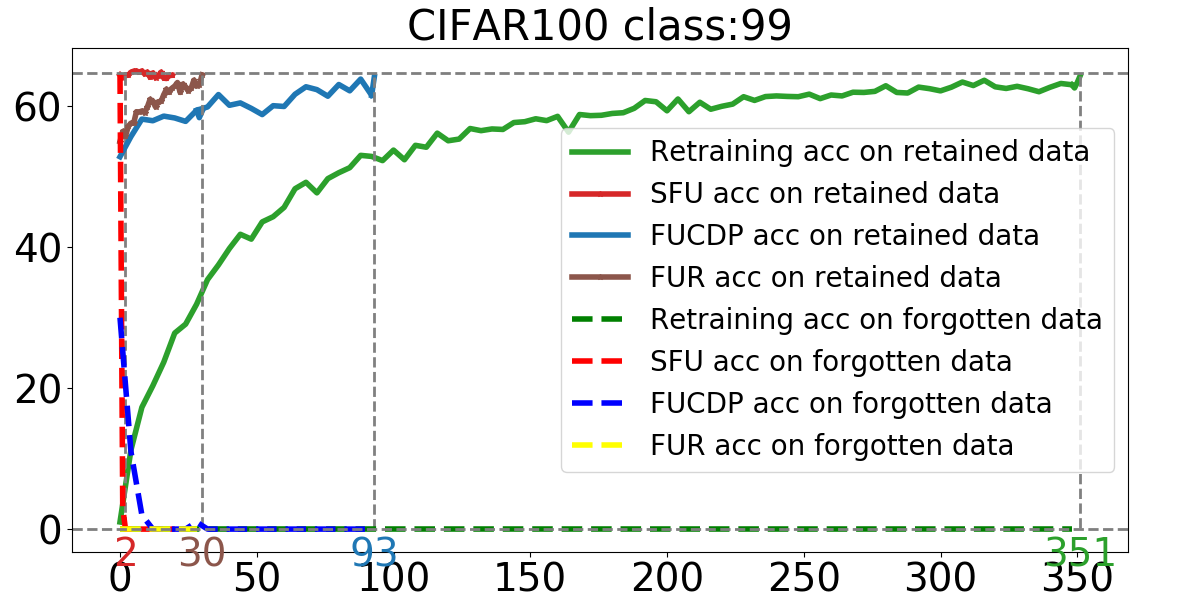}
                \vspace{0.75mm}
                \includegraphics[width=\linewidth]{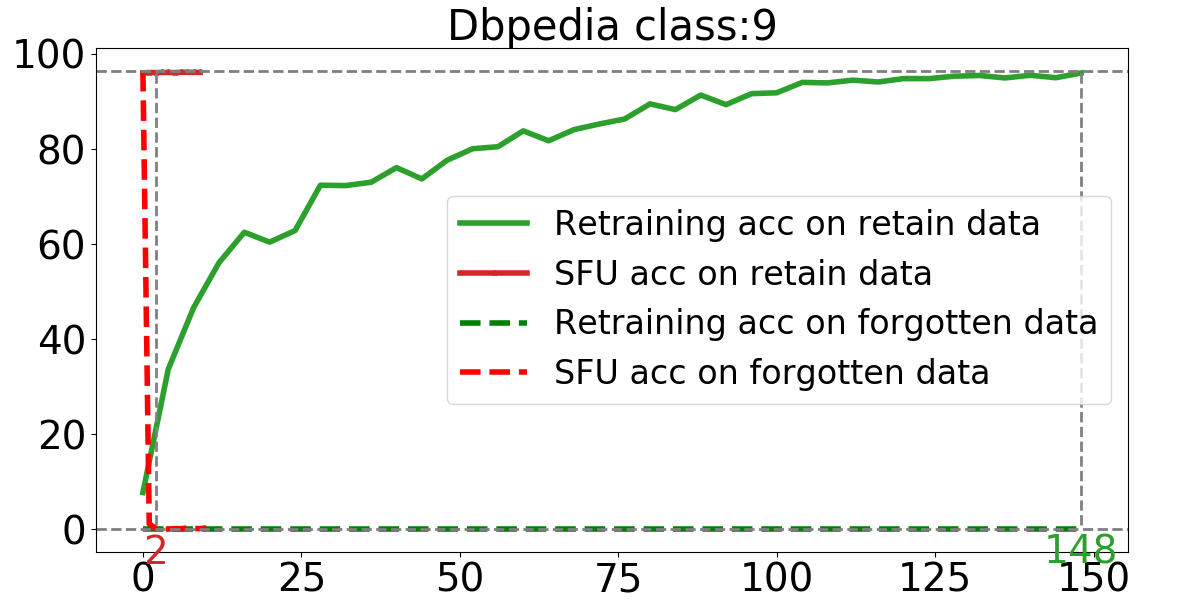}
                \centering Communication rounds
            \end{minipage}
        \end{minipage}
    \end{tabular}
    \caption{Test accuracy of models using different unlearning methods on forgotten and retained data. 
    Three different classes of data on each of the three different datasets are randomly selected to be unlearned using different methods. These data are randomly distributed on different clients according to non-iid. The lines of different colors represent the unlearning effects of different methods. The numbers of different colors on the horizontal axis represent the number of communication rounds required by each method to complete unlearning.}
    \label{fig3}
\end{figure*}

\subsubsection{Unlearning Efficacy}
We employ two metrics to evaluate unlearning efficacy: the accuracy of the model on forgotten data before and after unlearning, and the success rate of backdoor attacks.  
Firstly, we compare the prediction accuracy on forgotten data before and after unlearning across different FU methods. Theoretically, the more thoroughly the model unlearns, the closer this value approaches the prediction accuracy of retraining, which is nearly 0. As shown in Table \ref{tab:total}, we randomly select class data from each dataset for unlearning and report results for three specific classes. For the image datasets CIFAR-10 and CIFAR-100, SFU, FUCDP, and FUR achieve post-unlearning prediction accuracies that are nearly identical to the retraining baseline, approaching 0. On the additional text dataset Dbpedia, SFU similarly attains a prediction accuracy close to that of retraining, also approaching 0.  
From the perspective of prediction accuracy, these methods demonstrate unlearning efficacy comparable to retraining, effectively achieving comprehensive removal of forgotten data.  

To further validate unlearning efficacy, we employ backdoor attacks as an additional evaluation metric. Prior to unlearning, we implant backdoors into the forgotten data and then compare the backdoor attack success rate (BD\_ASR) before and after unlearning. Theoretically, BD\_ASR should be high before unlearning and significantly lower afterward. The lower the BD\_ASR after unlearning, the more effective the unlearning process is. 
As shown in Table \ref{tab:total}, using the baseline retraining for unlearning across three randomly selected classes in each dataset, BD\_ASR drops from high initial values to nearly 0. Similarly, after applying SFU for unlearning, BD\_ASR for the three classes in each dataset also declines to levels comparable to retraining, approaching 0. However, when FUCDP and FUR are employed, BD\_ASR decreases to approximately 2$\%$ on CIFAR-10 and CIFAR-100, which is slightly higher than the values achieved by retraining and SFU.  
From the perspective of BD\_ASR, SFU exhibits excellent unlearning efficacy, achieving a performance level similar to retraining. In contrast, FUCDP and FUR exhibit slightly weaker unlearning efficacy, falling short of SFU and retraining.

\subsubsection{Unlearning Fidelity}
We evaluate unlearning fidelity by examining the prediction accuracy of the model on retained data before and after unlearning. Ideally, an effective unlearning process should only affect forgotten data while preserving the model performance on retained data. As shown in Table \ref{tab:total}, when unlearning is performed using retraining, the prediction accuracy on retained data remains almost identical to its pre-unlearning accuracy. Similarly, after unlearning with SFU and FUCDP, the prediction accuracy on retained data remains close to its original level. However, for FUR, the results presented in the table show a noticeable decline in model performance after unlearning, which is non-negligible. This indicates that FUR negatively impacts retained data during the unlearning process. To address this issue, we apply additional fine-tuning to restore model performance. In contrast, FUCDP inherently includes a fine-tuning step, which allows it to recover performance effectively, as observed in the table.

The detailed process is illustrated in Figure \ref{fig3}, where the prediction accuracy on retained data changes over different training iterations. When unlearning is performed using retraining (green line), the accuracy gradually increases. FUCDP (blue line) starts with pruning and then uses fine-tuning to restore model performance. The figure shows the process of FUCDP fine-tuning, where the prediction accuracy on the retained data slowly increases. Similarly, FUR (brown line) exhibits a recovery trend after fine-tuning, eventually restoring prediction accuracy to pre-unlearning levels. For better visualization, the unlearning epochs for SFU are extended. The results clearly show that SFU (red line) maintains a stable prediction accuracy on retained data throughout the unlearning process, without any noticeable degradation. This suggests that SFU effectively preserves model performance during unlearning, eliminating the need for additional fine-tuning.

On the other hand, the prediction accuracy on the forgotten data rapidly decreases for all methods. The retraining approach (green dashed line) maintains an accuracy of 0, while FUCDP (blue dashed line) and FUR (yellow dashed line) stabilize near 0 after unlearning. SFU (red dashed line) also rapidly drops to 0 within very few iterations. This trend aligns with the conclusions from the previous section, confirming that all methods successfully remove information from forgotten data while exhibiting varying effects on retained data.
Additionally, the numbers on the x-axis in Figure \ref{fig3} represent the number of training epochs required for each FU method to complete unlearning. It is evident that the main factor influencing efficiency is the recovery of model performance. FUCDP and FUR require additional fine-tuning steps to restore accuracy, following a retraining-like process that demands more iterations. In contrast, SFU maintains stable model performance throughout unlearning, eliminating the need for fine-tuning and demonstrating superior efficiency and fidelity.

\subsubsection{Unlearnng Efficiency}
We use two metrics to measure the unlearning efficiency of these FU methods: number of communication rounds and speed-up ratio in time compared to retraining.
As shown in Table \ref{tab:efficiency}, in terms of communication rounds, the results across all datasets indicate that SFU, FUR, and FUCDP require significantly fewer communication rounds to complete unlearning than retraining. Among them, SFU demonstrates the greatest efficiency, requiring only 1-2 rounds on average, substantially reducing communication overhead and outperforming both FUR and FUCDP.

Regarding the speedup ratio, while FUCDP and FUR already achieve significant improvements over retraining, SFU consistently exhibits a much higher speedup across all datasets, highlighting its superior computational efficiency. This suggests that SFU not only ensures effective unlearning but also significantly reduces computational costs, making it a more efficient approach.
Additionally, we observe that the reduction in communication rounds and the improvement in speedup ratio are more pronounced on CIFAR-10 and DBpedia compared to CIFAR-100. This is likely because CIFAR-100 has a larger number of classes and higher task complexity, making unlearning more challenging. Similarly, while DBpedia is a text dataset, its higher information density increases processing difficulty compared to CIFAR-10, leading to variations in efficiency across datasets.
In conclusion, SFU significantly enhances unlearning efficiency for class data, far surpassing the baseline retraining method and outperforming the current SOTA methods FUCDP and FUR.

\begin{table*}[h!]
    \caption{Performance of SFU when unlearning multiple classes of data. We show the performance of SFU for the first-class data and the last-class data of unlearning, as well as the overall time speedup ratio and the number of communication rounds required.}
    \label{tab:multi}
    \centering
    \setlength{\tabcolsep}{2mm}
    \renewcommand{\arraystretch}{1.5}
    \begin{tabular}{ccccccccccccccc}
    \toprule[1.5pt] 
    \multicolumn{1}{c|}{\multirow{3}{*}{\begin{tabular}[c]{@{}c@{}}Id of \\ target class\end{tabular}}} & \multicolumn{1}{c|}{\multirow{3}{*}{\begin{tabular}[c]{@{}c@{}}Before \\ unlearning\end{tabular}}} & \multicolumn{6}{c|}{\begin{tabular}[c]{@{}c@{}}After unlearning\\ first class\end{tabular}} & \multicolumn{6}{c|}{\begin{tabular}[c]{@{}c@{}}After unlearning\\ last class\end{tabular}} & \multirow{3}{*}{\begin{tabular}[c]{@{}c@{}}Time \\ speed-up\end{tabular}} \\ \cline{3-14}
    \multicolumn{1}{c|}{} & \multicolumn{1}{c|}{} & \multicolumn{2}{c}{\begin{tabular}[c]{@{}c@{}}Acc on \\ forgotten data\end{tabular}} & \multicolumn{2}{c}{\begin{tabular}[c]{@{}c@{}}Acc on \\ retained data\end{tabular}} & \multicolumn{2}{c|}{\begin{tabular}[c]{@{}c@{}}Rounds\\ required\end{tabular}} & \multicolumn{2}{c}{\begin{tabular}[c]{@{}c@{}}Acc on \\ forgotten data\end{tabular}} & \multicolumn{2}{c}{\begin{tabular}[c]{@{}c@{}}Acc on \\ retained data\end{tabular}} & \multicolumn{2}{c|}{\begin{tabular}[c]{@{}c@{}}Rounds \\ required\end{tabular}} &  \\ \cline{3-14}
    \multicolumn{1}{c|}{} & \multicolumn{1}{c|}{} & Retrain & SFU & Retrain & SFU & Retrain & \multicolumn{1}{c|}{SFU} & Retrain & SFU & Retrain & SFU & Retrain & \multicolumn{1}{c|}{SFU} &  \\ \hline
    \multicolumn{15}{c}{CIFAR-10} \\ \hline
    \multicolumn{1}{c|}{0+5} & \multicolumn{1}{c|}{84.82} & 0.03 & 0.64 & 84.02 & 84.63 & 314 & \multicolumn{1}{c|}{1} & 0.24 & 0.35 & 83.46 & 84.02 & 302 & \multicolumn{1}{c|}{2} & 146.62x \\
    \multicolumn{1}{c|}{0+5+9} & \multicolumn{1}{c|}{84.93} & 0.12 & 0.92 & 85.32 & 83.43 & 304 & \multicolumn{1}{c|}{2} & 0.03 & 0.42 & 84.23 & 84.72 & 298 & \multicolumn{1}{c|}{1} & 135.35x \\ \hline
    \multicolumn{15}{c}{CIFAR-100} \\ \hline
    \multicolumn{1}{c|}{0+5} & \multicolumn{1}{c|}{64.34} & 0.04 & 0.36 & 64.14 & 63.45 & 377 & \multicolumn{1}{c|}{2} & 0.03 & 0.34 & 64.37 & 64.93 & 362 & \multicolumn{1}{c|}{2} & 42.32x \\
    \multicolumn{1}{c|}{0+5+9} & \multicolumn{1}{c|}{64.53} & 0.14 & 0.43 & 64.03 & 63.89 & 363 & \multicolumn{1}{c|}{2} & 0.07 & 0.51 & 64.87 & 63.99 & 348 & \multicolumn{1}{c|}{2} & 40.46x \\ \hline
    \multicolumn{15}{c}{Dbpedia} \\ \hline
    \multicolumn{1}{c|}{0+5} & \multicolumn{1}{c|}{96.43} & 0.04 & 0.06 & 96.45 & 95.98 & 148 & \multicolumn{1}{c|}{1} & 0.05 & 0.06 & 96.34 & 96.08 & 140 & \multicolumn{1}{c|}{1} & 42.50x \\
    \multicolumn{1}{c|}{0+5+9} & \multicolumn{1}{c|}{96.76} & 0.05 & 0.04 & 96.44 & 96.52 & 142 & \multicolumn{1}{c|}{2} & 0.06 & 0.04 & 96.43 & 96.67 & 135 & \multicolumn{1}{c|}{1} & 40.05x \\ 
    \bottomrule[1.5pt]
    \end{tabular}
\end{table*}

\subsection{Ablation Study}

\begin{figure*}[t]
    \centering
    \begin{tabular}{cc}
        \begin{minipage}[b]{0.02\linewidth}
            \centering
            \raisebox{0.4\height}{\rotatebox{90}{\text{Test accuracy( $\%$ )}}}
        \end{minipage}
        &
        \begin{minipage}[b]{0.98\linewidth}
            \centering
            \begin{minipage}[b]{0.32\linewidth}
                \includegraphics[width=\linewidth]{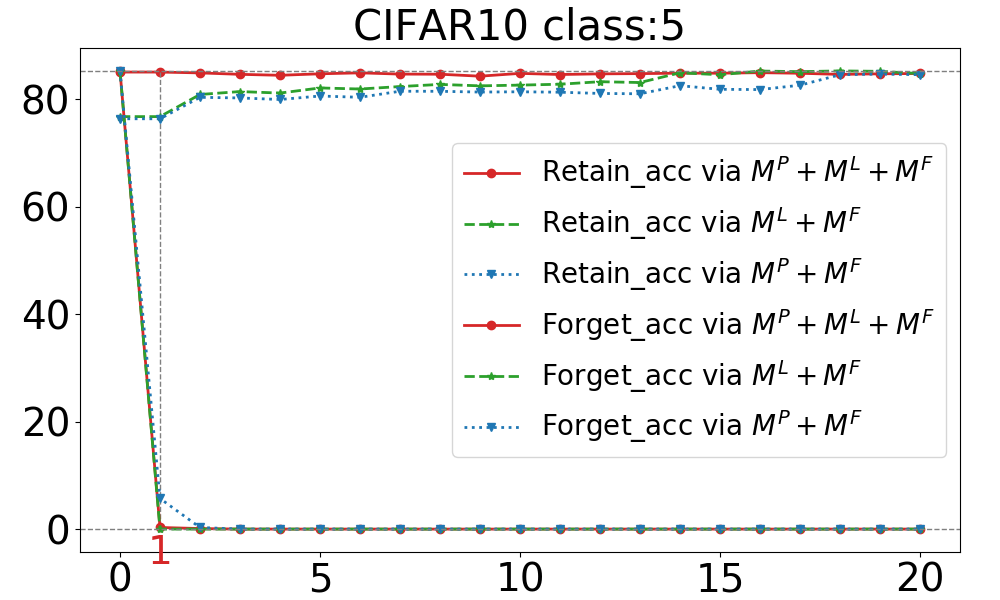}
                \centering Communication rounds
            \end{minipage}
            \begin{minipage}[b]{0.32\linewidth}
                \includegraphics[width=\linewidth]{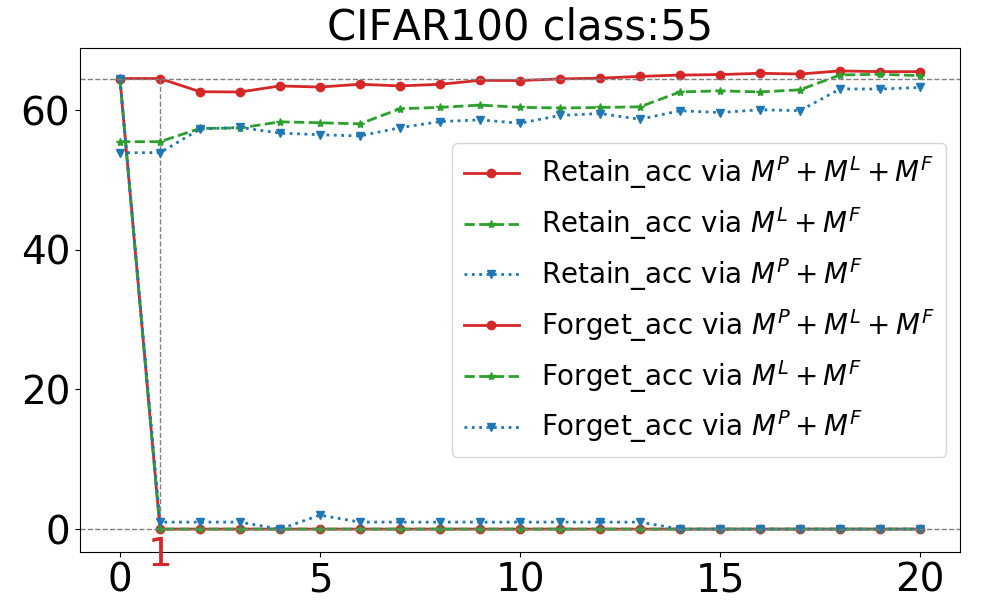}
                \centering Communication rounds
            \end{minipage}
            \begin{minipage}[b]{0.32\linewidth}
                \includegraphics[width=\linewidth]{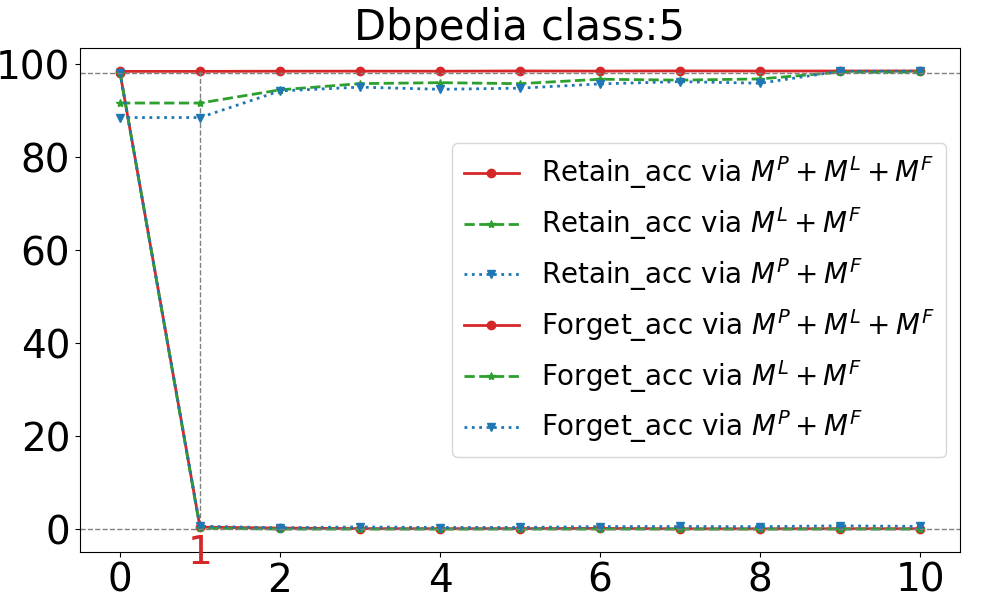}
                \centering Communication rounds
            \end{minipage}
        \end{minipage}
    \end{tabular}
    \caption{The unlearning performance of SFU under different combinations of teacher models. The lines of different colors represent the unlearning effects of different combinations of teacher models}
    \label{fig4}
\end{figure*}

\begin{figure*}[!h]
    \centering
    \begin{tabular}{cc}
        \begin{minipage}[b]{0.02\linewidth}
            \centering
            \raisebox{0.4\height}{\rotatebox{90}{\text{Test accuracy( $\%$ )}}}
        \end{minipage}
        &
        \begin{minipage}[b]{0.98\linewidth}
            \centering
            \begin{minipage}[b]{0.32\linewidth}
                \includegraphics[width=\linewidth]{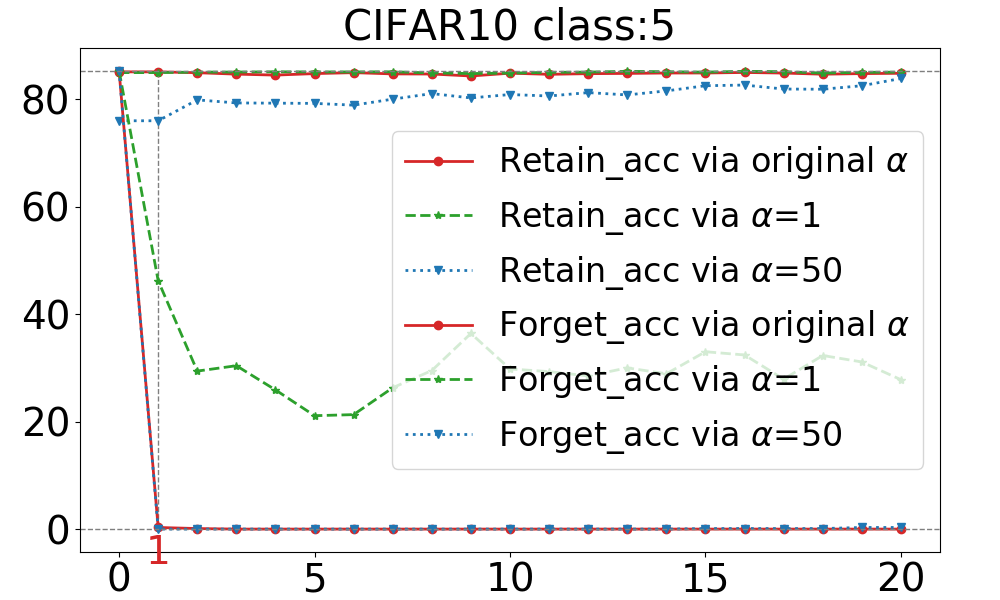}
                \centering Communication rounds
            \end{minipage}
            \begin{minipage}[b]{0.32\linewidth}
                \includegraphics[width=\linewidth]{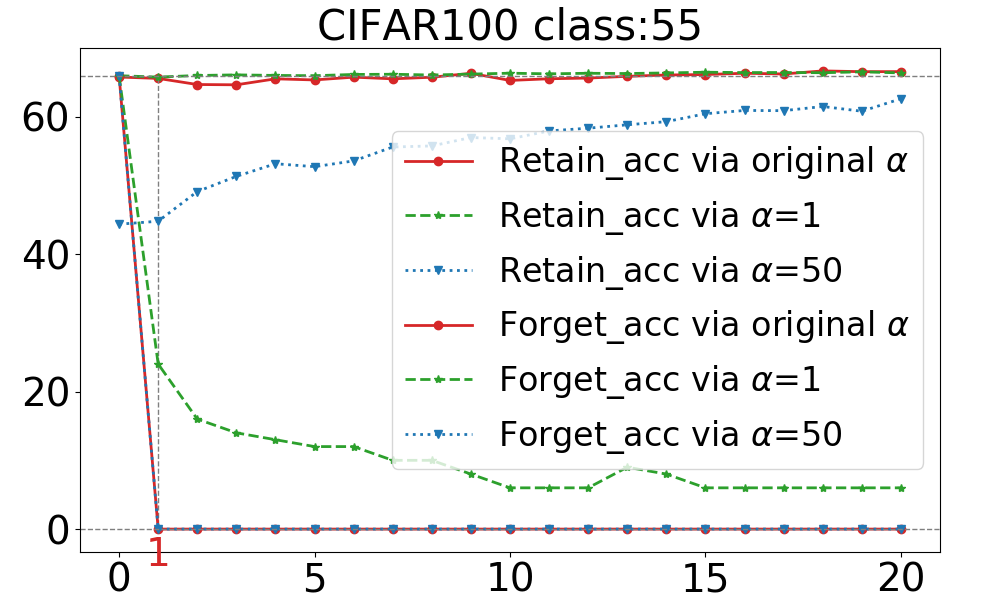}
                \centering Communication rounds
            \end{minipage}
            \begin{minipage}[b]{0.32\linewidth}
                \includegraphics[width=\linewidth]{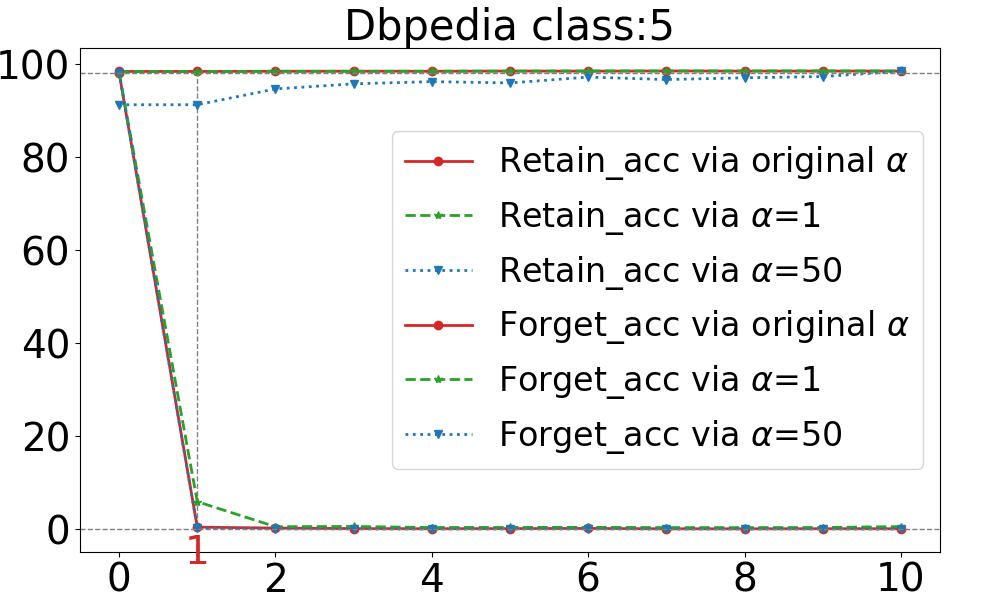}
                \centering Communication rounds
            \end{minipage}
        \end{minipage}
    \end{tabular}
    \caption{Comparison of the unlearning performance of SFU with different values of hyperparameters \( \alpha \).
    We set three different orders of magnitude of \( \alpha \) to compare its impact on unlearning: \( \alpha \) calculated according to the proposed method and \( \alpha \) one order of magnitude above and below it. We select one class of data from each of the three datasets for unlearning.
    }
    \label{fig5}
\end{figure*}

\subsubsection{Combinations of Teachers Models}
We develop a multi-teacher system comprising three distinct teacher models: the performance preservation teacher model (\(M^P\)), the forgetting teacher model (\(M^F\)), and the label-based preservation teacher model (\(M_i^L\)). Based on their roles, \(M^F\) is responsible for forgetting, while \(M^P\) and \(M^L\) focus on performance preservation. To validate the rationale behind this design and assess the necessity of each teacher model, we conduct extensive ablation experiments. Specifically, we perform three sets of comparative experiments: using only the \(M^P\) and \(M^F\) combination, only the \(M^L\) and \(M^F\) combination, and all three models (\(M^P\), \(M^L\), and \(M^F\)) together. 
As shown in Figure \ref{fig4}, all three combinations effectively remove the influence of forgotten data on three datasets. However, in terms of performance preservation, the combinations of \(M^P\) with \(M^F\) or \(M^L\) with \(M^F\) led to some performance degradation. Only the complete combination of \(M^P\), \(M^L\), and \(M^F\) successfully remove the influence of forgotten data while maintaining model performance, thus significantly enhancing the efficiency of unlearning.
In summary, each teacher model in our multi-teacher system serves a unique and essential role, making them all indispensable to the overall effectiveness of the system.

\subsubsection{Hyperparameters}
\label{sec:ex_hyper}
Given the varying amounts of target data and residual data across different datasets and unlearning requests, we introduce a hyperparameter in the loss function to balance these two aspects. From the perspective of the teacher models, we aim for a balance between ``knowledge of forgetting" and ``knowledge of performance" when guiding the student model. This balance is crucial to effectively removing the influence of the target data without compromising overall model performance.

To accommodate different datasets, we define \( \alpha = {num(x^r)} / {num(x^f)} \), where \( x^r \) represents the retained data and \( x^f \) the target data to be forgotten. We conduct extensive experiments to validate the effectiveness of this parameter across various values of \( \alpha \). The results, presented in Figure \ref{fig5}, indicate that the choice of \( \alpha \) significantly influences the balance between performance retention and influence removal. For instance, in the CIFAR-10 dataset when unlearning class 5, we compare the default \( \alpha = 9 \) (as defined) with \( \alpha = 1 \) and \( \alpha = 50 \). The results show that \( \alpha = 1 \) preserves model performance but fails to effectively remove the influence of the target data, while \( \alpha = 50 \) successfully removes the influence but at the cost of decreased performance. The default \( \alpha \) achieves a balance between these extremes, maintaining performance while effectively removing the influence of forgotten data. Similar observations were made for CIFAR-100 and the text dataset Dbpedia.
In conclusion, when \( \alpha \) is too small, the model retains performance but fails to remove the influence of the forgotten data effectively. Conversely, a large \( \alpha \) results in effective influence removal but degrades performance. Our definition of \( \alpha \) thus strikes an optimal balance, ensuring that the model removes the influence of forgotten data without compromising its performance and is adaptable to various datasets and unlearning requests.

\subsubsection{Multiple classes unlearning}
For multiple classes unlearning, SFU only needs to loop through the execution after completing a single class. The results are shown in Table \ref{tab:multi}, where we run multiple classes unlearning experiments on image and text datasets, respectively. It is obvious that the efficiency of SFU unlearning is about the same as that of single-class unlearning. At the completion of the last class unlearning, the accuracy of the model on the retained data is still very close to the state before unlearning, which also shows that the model performance is well preserved. Moreover, at the completion of unlearning for each class, the accuracy on the forgotten data was close to 0, similar to that of retraining, which also indicates that the model did a good job of removing the influence of the target data. Most importantly, for each class that needs to be forgotten, SFU only needs 1 or 2 rounds of communication to complete the unlearning, which greatly improves communication efficiency and time efficiency.

\section{Conclusion}
\label{sec:conclusion}
In this study, we propose a streamlined federated unlearning method called SFU that significantly enhances unlearning efficiency by integrating the steps of influence removal and performance recovery. We design a multi-teacher system where three distinct teacher models guide the unlearned model to remove the influence of target data while maintaining model performance. SFU does not require additional computational or storage resources. It adheres to the security requirements of FL without necessitating global data access and is not constrained by data distribution. Furthermore, SFU exhibits great flexibility, seamlessly integrating into any training stage of FL. Experimental results demonstrate that SFU is highly generalizable, applicable to various models, and effective on both text and image datasets. Compared to retraining, SFU significantly improves time and communication efficiency and substantially outperforms existing SOTA methods. Additionally, the effectiveness of SFU is validated through backdoor attacks. In the future, we plan to explore more efficient and generalized FU methods for more unlearning scenarios and goals.


\ifCLASSOPTIONcaptionsoff
  \newpage
\fi

\bibliographystyle{IEEEtran}         
\bibliography{sfu_reference}


\end{document}